\documentclass[10pt, twocolumn, journal, web]{article}

\usepackage[ruled,vlined]{algorithm2e}
\usepackage{times}
\usepackage[margin=16mm]{geometry}
\usepackage{amsmath,amssymb,amsfonts}

\usepackage[utf8]{inputenc}
\usepackage{utfsym}
\usepackage{multirow}
\usepackage[normalem]{ulem}
\usepackage{amsmath,amssymb,amsfonts}
\usepackage{algorithmic}
\useunder{\uline}{\ul}{}
\usepackage{url}
\def\BibTeX{{\rm B\kern-.05em{\sc i\kern-.025em b}\kern-.08em
    T\kern-.1667em\lower.7ex\hbox{E}\kern-.125emX}}

\usepackage{graphicx}
\usepackage{authblk}
\usepackage{booktabs}
\usepackage{array}

\usepackage{url}
\usepackage{verbatim}
\usepackage{graphicx}
\usepackage{pifont} 
\usepackage{subfigure}
\usepackage{rotating}
\usepackage{bbm}
\usepackage{utfsym}
\usepackage{multirow}
\usepackage[normalem]{ulem}
\useunder{\uline}{\ul}{}
\usepackage{amsthm,amssymb}
\usepackage{mathrsfs}
\usepackage[colorlinks=true, linkcolor=blue, citecolor=blue, urlcolor=blue]{hyperref}

\usepackage{caption}
\usepackage{amsmath}
\usepackage{rotating}
\usepackage{bbm}
\usepackage{utfsym}
\usepackage{multirow}
\usepackage[normalem]{ulem}
\useunder{\uline}{\ul}{}
\usepackage{pifont} 
\usepackage{setspace}

\usepackage[colorlinks=true, linkcolor=blue, citecolor=blue, urlcolor=blue]{hyperref}

\newcommand{\xvect}{\mbox{\bf x}}

\newcommand{\Avect}{\mbox{\bf A}}

\newcommand{\Fvect}{\mbox{\bf F}}

\newcommand{\Kvect}{\mbox{\bf K}}

\newcommand{\Xvect}{\mbox{\bf X}}

\makeatletter
\renewcommand\AB@affilsepx{, \protect\Affilfont}
\makeatother
\usepackage{tabularx}
\providecommand{\keywords}[1]
{
  \small	
  \textbf{\textit{Keywords---}} #1
}

\begin{document}

\title{\textbf{MCFCN: Multi-View Clustering via a Fusion-Consensus Graph Convolutional Network}}
\author[1]{Chenping Pei}
\author[1, 2]{Fadi Dornaika \thanks{Corresponding author}}
\author[3]{Jingjun Bi}
\affil[1]{\textit{University of the Basque Country}}
\affil[2]{\textit{IKERBASQUE}}
\affil[3]{\textit{North China University of Water Resources and Electric Power}}

\affil[ ]{

\small\texttt{chen$\_$ping.pei@foxmail.com, fadi.dornaika@ehu.eus, bi.jingjun@outlook.com}}
\date{}
\maketitle

\date{}
\maketitle
\begin{abstract}
Existing Multi-view Clustering (MVC) methods based on subspace learning focus on consensus representation learning while neglecting the inherent topological structure of data. Despite the integration of Graph Neural Networks (GNNs) into MVC, their input graph structures remain susceptible to noise interference. Methods based on Multi-view Graph Refinement (MGRC) also have limitations such as insufficient consideration of cross-view consistency, difficulty in handling hard-to-distinguish samples in the feature space, and disjointed optimization processes caused by graph construction algorithms. To address these issues, a  Multi-View Clustering method via a Fusion-Consensus Graph Convolutional Network (MCFCN) is proposed. The network learns the consensus graph of multi-view data in an end-to-end manner and learns effective consensus representations through a view feature fusion model and a Unified Graph Structure Adapter (UGA). It designs Similarity Matrix Alignment Loss (SMAL) and Feature Representation Alignment Loss (FRAL). With the guidance of consensus, it optimizes view-specific graphs, preserves cross-view topological consistency, promotes the construction of intra-class edges, and realizes effective consensus representation learning with the help of GCN to improve clustering performance. MCFCN demonstrates state-of-the-art performance on eight multi-view benchmark datasets, and its effectiveness is verified by extensive qualitative and quantitative implementations. The code will be provided at \textcolor{blue}{\url{https://github.com/texttao/MCFCN}}.
\end{abstract}

 \hspace{10pt}

\section{Introduction}
In recent years, with the increasing advancement of multimodal data acquisition technologies, multi-view data has shown a trend of explosive growth. This type of data contains rich and diverse information dimensions and can provide strong support for solving complex tasks \cite{ma2024multilevel,zhang2021cmc}. In autonomous driving applications \cite{man2023bev,li2023mseg3d}, the converged installation of high-resolution imaging cameras, millimeter-wave radar systems, and ultrasonic ranging sensors has substantially upgraded the vehicle's environmental perception in intricate traffic conditions. In multimedia analysis, researchers can describe a single image not only via multiple feature encoding techniques but also by integrating text description information. MVC \cite{li2018survey,fang2023comprehensive,wen2022survey,zou2024revisiting,liu2024learning,liu2024progressive}, a core subfield within the multi-view learning paradigm, has attracted extensive attention in academia and industry recently. Its research emphasis lies in mining complementary information across multi-view data to enable efficient cluster partitioning of samples \cite{Fang2023}.

The powerful capabilities demonstrated by deep neural networks in representation learning tasks have promoted the emergence of a batch of MVC methods based on subspace learning \cite{trosten2023effects,yan2023gcfagg,li2019deep,chen2020multi,trosten2021reconsidering}. However, such methods often focus on the learning process of consensus representation but neglect the inherent topological structure of the data itself that is naturally suitable for clustering tasks. Leveraging their advantages in geometric structure mining and node representation learning, Graph Neural Networks (GNNs) have been incorporated into the MVC method system \cite{lin2024dual,xiao2023dual}. Although certain progress has been made, existing methods have deficiencies in controlling the quality of the input graph structure. Most cutting-edge methods directly construct graph structures based on raw data, which are prone to being interfered by noise \cite{xiao2023dual,wang2021consistent}. This, in turn, introduces additional noise into the GNN training process, leading to the deterioration of clustering performance.

To address the above challenges, a series of clustering methods based on MGRC have been proposed \cite{wang2024surer,du2023robust}. This approach first constructs initial graphs from the raw features of each view, and then learns more effective feature representations to optimize the graph structure of each view. While these methods have achieved moderate effectiveness in enhancing the quality of view-specific graphs, notable limitations still persist: First, they only focus on optimizing view-specific graphs using intra-view information and do not fully consider the maintenance of cross-view structural consistency, resulting in impaired integrity of consensus representation learning. Therefore, constructing a mechanism for topological structure alignment across different views and consensus graph structure learning is of great importance. Second, constructing the underlying graph structure solely based on node similarity of raw features fails to effectively handle samples from different clusters in high-dimensional feature spaces. Meanwhile, incorporating semantic-level information is expected to enhance the effectiveness of optimal graph learning. Third, graphs are mostly constructed via $k$ highest similar samples (kNN) in existing methods, with the kNN sorting algorithm being non-differentiable in deep neural network contexts. This constraint compels existing GNN-based approaches to operate in a two-stage manner, leading to a disjointed optimization process.

To address this challenge, as illustrated in Figure \ref{MFGCC_all}, we introduce a Multi-View Fusion Consensus Graph Convolutional Network for multi-view clustering tasks. It aims to obtain the consensus graph of multi-view data in an end-to-end manner and learn effective consensus representations.Specifically, we design a multi-view feature fusion model and a Unified Graph Structure Adapter (UGA). Initially, we perform effective fusion of multi-view features. By promoting intra-class connections in the view-specific structure graph, we implicitly refine the graph topology while maintaining cross-view structural consistency. Concurrently, drawing inspiration from graph structure learning, we leverage the Graph Convolutional Network(GCN) framework to derive a consensus representation of the features. We innovatively introduce two loss functions, namely Similarity Matrix Alignment Loss (SMAL) and Feature Representation Alignment Loss (FRAL), to build a unified training framework to jointly discover consensus topology and representation information while generating high-quality pseudo-labels through self-supervised learning.

In summary, our research contributions can be categorized into three main aspects:
\begin{itemize}
\item
We devise the MCFCN Algorithm for multi-view clustering tasks. Leveraging the View Feature Fusion Module and Unified Graph Structure Adapter, MCFCN conducts joint learning and iterative optimization of view-specific graph topologies and representations.

\item
We formulate two specialized loss functions-SMAL and FRAL. These losses serve as guiding metrics, facilitating the optimization of view-specific graphs via consensus-driven mechanisms and ensuring the preservation of cross-view topological coherence.

\item
The proposed method demonstrates state-of-the-art performance on eight real-world multi-view benchmark datasets. Extensive qualitative and quantitative experiments fully validate the superior clustering effectiveness of MCFCN.
\end{itemize}
This paper is structured as follows. Section \ref{sec:related} reviews related  techniques and recent developments. Section \ref{sec:method} introduces the core concepts of MCFCN.  Section \ref{sec:setup} describes the experimental setup. Section \ref{sec:Performance} reports and analyzes the experimental findings. Section  \ref{sec:conclusion} concludes the paper.

\section{Related Work}
\label{sec:related}
This section first overviews MVC research, then introduces graph-based multi-view clustering methods and their advances.

\subsection{Main Methods of Multi-view Clustering} 
Multi-view clustering aims to improve clustering performance through the complementary information among different views. In recent years, with the emergence of numerous MVC methods, they can be roughly classified into the following categories according to their technical implementations: methods based on multi-kernel learning, methods based on co-learning, methods based on subspace learning, and methods based on graphs. Generally, multi-kernel learning methods construct and fuse the base kernels of different views to obtain a consistent kernel matrix that integrates multi-view information \cite{zhou2021multiple,liu2021incomplete,liu2020cluster,zhang2022multiple,wang2021late}. Co-learning methods guide the clustering process to maximize the consensus information and make the clustering results of each view tend to be consistent. Subspace learning methods are the mainstream methods in MVC research, aiming to find a shared representation space for each view while retaining the unique distribution information of each view as much as possible \cite{huang2022learning}. Some approaches leverage matrix factorization or data self-representation properties to accomplish shallow feature representation learning. Recently, propelled by the formidable modeling capabilities of deep learning for non-linear and intricate data, MVC methods based on deep multi-view subspaces  \cite{xu2022multi,wang2023triple,trosten2023effects,li2019deep,trosten2021reconsidering} have attracted escalating attention. For instance, Xu et al.  \cite{xu2022multi} incorporated multi-level contrastive learning into MVC to capture features across diverse levels, including low-level, high-level, and semantic-level features. Despite their notable accomplishments, these methods inadvertently overlook the geometric structure inherent in the data, thereby restricting their performance.
\subsection{Graph-based Multi-view Clustering} 
Compared with other methods, graph-based methods focus on considering the geometric structure information within each view \cite{huang2023self,cheng2021multi,wen2021structural,ling2023dual,peng2019comic,peng2019comic,pan2021multi,xiao2023dual}. In this research branch, early approaches primarily learned the consensus graph by applying diverse regularization terms to a specific view, followed by generating clustering outcomes via algorithms like spectral clustering \cite{wang2019gmc,li2021consensus,zhan2018multiview,Yu2025}. In recent years, with the wide application of deep learning, fully exploring the structural information of multi-view data through neural networks has become an important research method. Xia et al. (2022) leveraged clustering labels to guide the feature representation learned by a multi-view shared graph attention encoder module \cite{xia2022multi}. Huang et al. (2023) put forward a self-supervised graph attention network designed specifically for deep weighted multi-view clustering \cite{huang2023self}. These Graph-based Deep Multi-view Clustering Methods have improved clustering performance to a certain extent, but still have the problems mentioned above, that is, the pre-constructed graph is affected by the noise of the original data, and it is difficult to propagate structural relationships across views.

\begin{table*}[htbp]
\caption{Main notations used in the paper.}
\label{Notations_table}
\centering
\begin{tabular}{ll}
\hline
Notation     & Description  \\ \hline
$N$ & Number of data samples\\
$C$ & Number of clusters \\
$d$ & Dimension of the view features after projection \\

$\Xvect^v= [\xvect_1; \ \xvect_2; \ldots; \xvect_n]\in \mathbb{R}^{N\times d_v}$ & The data matrix in $v$-th view\\
$d_v$ & Dimensionality of data in the $v$-th view\\

$\Fvect_f\in \mathbb{R}^{N\times (d\times v)}$ &Fused feature matrix \\
$\Avect_f\in \mathbb{R}^{N\times N}$ &Unified adjacency matrix \\
$\Kvect_f\in \mathbb{R}^{N\times N}$ &The kernel matrix of the fused features \\
$\Kvect^v\in \mathbb{R}^{N\times N}$ &The kernel matrices of each view in the original features \\
 \hline
\end{tabular}

\end{table*}

\begin{figure*}[h!]
  \centering
  \subfigure[Model architecture and loss function.]{\includegraphics[width=1\textwidth]{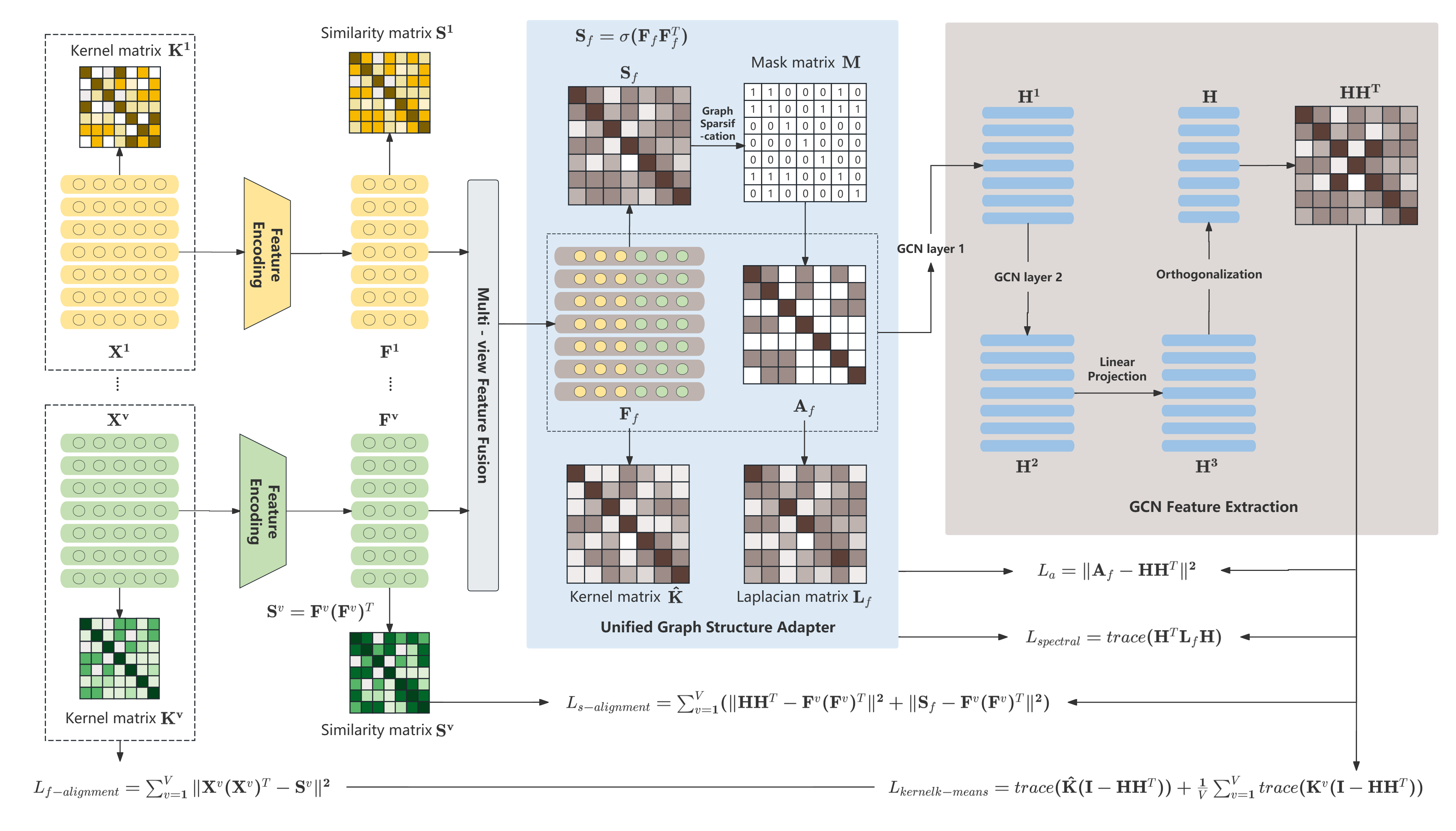}}\\
  \subfigure[Clustering.]{\includegraphics[width=0.8\textwidth]{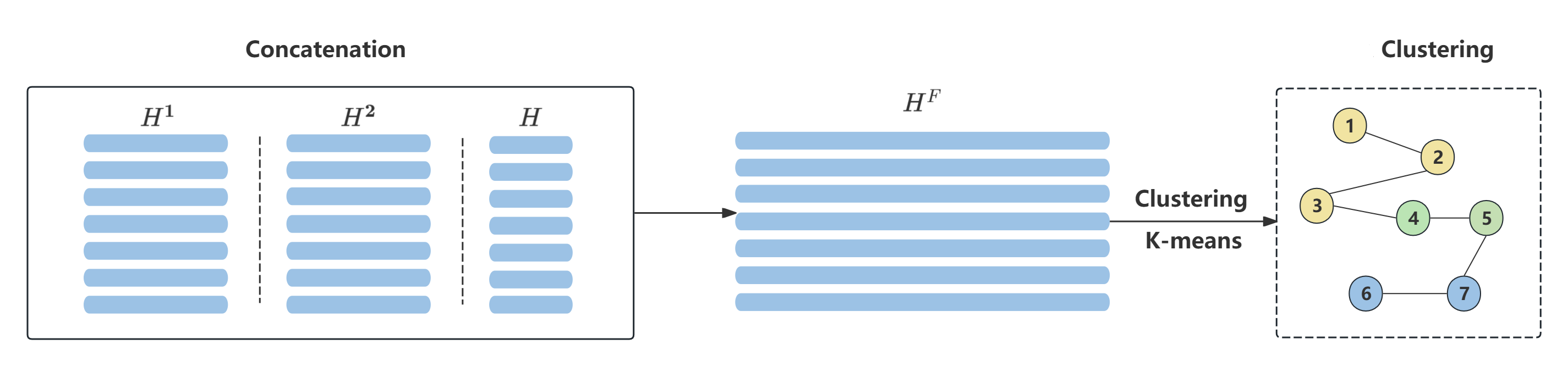}}
  \caption{The overall architecture of MFGCC. First, we perform linear transformation on each view, unify the feature dimensions, and conduct feature fusion. Additionally, a Unified Graph Structure Adapter is introduced to generate learnable graphs for the subsequent GCN, thereby facilitating the joint optimization of graph structures and their corresponding representations. After that, we extract features through the GCN to obtain the consensus vector representation, concatenate it with the output of the intermediate layer of the GCN, and then use the K-means algorithm to get the final clustering results. In addition, we design a loss function to guide the model to learn the view-consistent topological structure and its corresponding feature representations, so as to improve the model performance.}
\label{MFGCC_all}
\end{figure*}

\section{Multi-View Fusion Consensus Graph Convolutional Clustering Network}
\label{sec:method}
In this section, we propose a Multi-View Fusion Consensus Graph Convolutional Clustering Net-
work, named MCFCN (Multi-View Clustering via a Fusion-Consensus Graph Convolutional Network Algorithm). Given a multi-view dataset $\{\mathbf{X}^v\}_{v = 1}^{V} \in \mathbb{R}^{N\times d_v}$, where $V$ represents the number of views and $N$ is the number of samples, and $d_v$ is the dimension of the $v$-th view. MCFCN divides the samples into $C$ disjoint clusters by effectively fusing the features of each view. MCFCN is designed as an end-to-end optimization framework, which can effectively improve the clustering performance by mining the unified graph topology of each view. The overall architecture is shown in Figure \ref{MFGCC_all}.

In Section 3.1, we first introduced the notation used throughout the paper. In Section 3.2, the Multi-view Feature Fusion Module is introduced in detail to achieve the fusion of features between views. In Section 3.3 , the Unified Graph Structure Adapter and how to obtain the unified graph topology of each view are introduced in detail. In Section 3.5, we detail the GCN-based feature extraction network. In Section 3.6, we present the relevant loss functions of the model, including the clustering loss, graph structure loss, and feature similarity loss, and finally give the overall loss function of the model.

\subsection{Notations}
Throughout this work, scalars are symbolized by lowercase letters, while vectors are denoted with bold lowercase letters. Bold uppercase letters serve to represent matrices. Table \ref{Notations_table} presents a summary of the key notations used in this chapter.

\subsection{Multi-view Feature Fusion Module}
The objective of multi-view feature processing is to efficiently leverage the complementary feature data across views, enhance the salience of unified features, and mitigate the influence of noise. Therefore, fusing each view into a unified feature matrix is an effective approach. Since the feature dimensions and value ranges of each view vary, it poses challenges to the feature fusion process.

In the model, we first embed the original features of each view into a latent space of a unified dimension through linear transformation. Then, we perform column-wise L2 normalization on the embedded features to constrain the value range of the features, thereby reducing the negative impact of extreme values on the model. The specific formula is:
\begin{equation}
\mathbf{F}^v = norm(\mathbf{U}^v\mathbf{X}^v)
\label{eq:1}
\end{equation}
Finally, we fuse the multi-view features through a concatenation operation to obtain a unified feature representation $\mathbf{F}_{f}$. The formula is as follows:
\begin{equation}
\mathbf{F}_{f} = [\mathbf{F}^0, \mathbf{F}^1, \ldots, \mathbf{F}^V]
\label{eq:2}
\end{equation}

\subsection{Unified Graph Structure Adapter (UGA)}
Applying graph structures to multi-view data can effectively reflect the intrinsic topology of the data. By representing data samples as nodes and the relationships between samples as edges, the intrinsic topological structure of the data can be reflected. In multi-view clustering, corresponding graph structures can be constructed for data from different views to help capture the associations of data in different dimensions. For example, a graph constructed based on node similarity can present the proximity relationships between samples and provide basic information on data distribution for clustering.

In previous similar algorithms, the construction of graph structures often relies on the k highest similar samples (kNN) algorithm applied to the raw features of each view. Although this method can effectively obtain graph structures measured by distance or similarity between nodes, it is highly susceptible to noise. Moreover, the graph structures obtained from each view often vary greatly, which interferes with the model learning and clustering results.

In multi-view data, each view represents a different perspective of the same object. Therefore, we can propose the following hypotheses: 1. The graph structures of each view should be roughly the same. 2. Fusing multiple views can lead to more comprehensive and effective feature representations.

Based on these two hypotheses, we propose the unified graph structure adapter module. First, we calculate the fused similarity matrix $\mathbf{S}_{f}$ using the fused features $\mathbf{F}_{f}$. The formula is:
\begin{equation}
\mathbf{S}_{f} = \sigma(\mathbf{F}_{f}\mathbf{F}_{f}^T)
\label{eq:3}
\end{equation}
where $\sigma(\cdot)$ denotes the ReLU activation function. Although $\mathbf{S}_{f}$ can represent the relationships between each sample, it is a dense matrix, which contain many spurious and non relevant edges. In the meantime, current similar approaches commonly utilize the kNN algorithm for adjacency matrix acquisition. However, due to the sparsification of the sorting algorithm, the constructed graph is non-differentiable, resulting in a discontinuous optimization process and performance degradation.

In our method, the initial matrix is converted to   a mask (binary matrix $\mathbf{M} \in \mathbb{B}^{N\times N}$. Perform graph sparsification on \(\mathbf{S}_{f}\), When a value in $\mathbf{S}_{f}$ belongs to the k highest values in the associated row, the corresponding $M_{ij} = 1$; otherwise, $M_{ij} = 0$. The sparse similarity matrix is computed:
\begin{equation}
\dot{\mathbf{S}_{f}} = \mathbf{S}_{f} \odot \mathbf{M}
\label{eq:4}
\end{equation}
Considering that graph structures in real life are mostly symmetric, we further post-process $\dot{\mathbf{S}}_{f}$ to obtain the final fused adjacency matrix:
\begin{equation}
\mathbf{A}_{f} = \frac{\dot{\mathbf{S}}_{f}+\dot{\mathbf{S}}_{f}^T}{2}
\label{eq:5}
\end{equation}

\subsection{GCN Feature Extraction Network}
GCN can effectively capture the structural information in the original data and has become the de-facto architecture for many graph-based unsupervised and semi-supervised tasks. In our work, we adopt a three-layer GCN network as the feature extractor. Similar to other GCN-based methods, it consists of two GCN layers and a linear layer.

Previously, we have obtained the unified feature representation and unified graph structure of multi-view data. Here, we feed  $\mathbf{F}_{f}$ and $\mathbf{A}_{f}$ into the GCN to process them. The output of the first layer $\mathbf{H}^1 \in \mathbb{R}^{N\times h_1}$ and the output of the second layer $\mathbf{H}^2 \in \mathbb{R}^{N\times h_2}$ are calculated as follows:
\begin{equation}
\mathbf{H}^1 = \sigma(\hat{\mathbf{A}}_{f}\mathbf{F}_{f}\mathbf{W}^1)
\label{eq:6}
\end{equation}
\begin{equation}
\mathbf{H}^2 = \sigma(\hat{\mathbf{A}}_{f}\mathbf{H}^1\mathbf{W}^2)
\label{eq:7}
\end{equation}

The inner layers contain trainable matrices \(\mathbf{W}^1 \in \mathbb{R}^{(d\times v)\times h_1}\) and \(\mathbf{W}^2 \in \mathbb{R}^{h_1\times h^2}\) respectively, with \(h_1\) and \(h_2\) being the feature counts of hidden layers. The graph's normalized adjacency matrix, \(\hat{\mathbf{A}}_{f}\), is calculated as follows:

\begin{equation}
\hat{\mathbf{A}}_{f}=\hat{\mathbf{D}}_{f}^{-\frac{1}{2}}(\mathbf{A}_{f} + \mathbf{I})\hat{\mathbf{D}}_{f}^{-\frac{1}{2}}
\label{eq:8}
\end{equation}
where $\hat{\mathbf{D}}_{f}$ is a diagonal matrix defined as:
\begin{equation}
\hat{\mathbf{D}}_{f_{ij}}=\sum_{j}(\mathbf{A}_{f}+\mathbf{I})_{ij}
\label{eq:9}
\end{equation}
In this context, $\mathbf{I}$ signifies the identity matrix. Within the output layer (linear layer), we map $\mathbf{H}^2$ to a $C$-dimensional space, with $C$ denoting the cluster number.

\begin{equation}
\mathbf{H}^3=\mathbf{H}^2\mathbf{W}^3
\label{eq:10}
\end{equation}

\(\mathbf{W}^3 \in \mathbb{R}^{h_2\times C}\) is another learnable transformation matrix. Given that \(\mathbf{H}^3\) functions as the node representation, it needs to be orthogonal. Therefore, we apply Cholesky decomposition to \((\mathbf{H}^3)^T\mathbf{H}^3\):

\begin{equation}
(\mathbf{H}^3)^T\mathbf{H}^3 +\epsilon \cdot \mathbf{I}=\mathbf{Q}^T\mathbf{Q}
\label{eq:11}
\end{equation}
where $\mathbf{Q} \in \mathbb{R}^{C\times C}$ is a lower-triangular matrix. The orthogonal form of $\mathbf{H}^3$ is:
\begin{equation}
\mathbf{H}=\mathbf{H}^3(\mathbf{Q}^{-1})^T
\label{eq:12}
\end{equation}

\subsection{Unsupervised Learning Loss}
Our aim is to utilize GCN in an unsupervised manner to partition graph nodes into $C$ different clusters. To obtain more reliable clustering results, it is necessary to fuse and align the information from different views. The following parts mainly introduce our loss function.

\subsubsection{Multi-view Deep Kernel K-means Loss}
The original dataset contains $V$ views, with $V$ kernel functions. Meanwhile, through the feature fusion module, we obtain a unified feature representation which also has a kernel function. Our goal is to reasonably apply the above $V + 1$ kernel functions and minimize the following objective to achieve soft cluster assignment:
\begin{equation}
\begin{split}
L_{kernel k-means}=\sum_{j = 1}^{C}\sum_{\hat{x}_i\in C_j}\|\phi(\hat{x}_i)-\hat{m}_j\|^2   \\ +\frac{1}{V}\sum_{v = 1}^{V}\sum_{j = 1}^{C}\sum_{x_i^v\in C_j}\|\phi(x_i^v)-m_j^v\|^2
\label{eq:13}
\end{split}
\end{equation}
where $C$ is the number of clusters, $C_j$ is the $j$-th cluster, $\phi(\hat{x}_i)$ is a non-linear function of the unified feature representation $\hat{x}_i$, and $m_j^v$ and $\hat{m}_j$ are the $j$-th cluster centers of each view and the fused view respectively, given by the following equations:
\begin{equation}
m_j^v=\frac{1}{n_j^v}\sum_{x_i^v\in C_j}\phi(x_i^v)
\label{eq:14}
\end{equation}
\begin{equation}
\hat{m}_j=\frac{1}{\hat{n}_j}\sum_{\hat{x}_i\in C_j}\phi(\hat{x}_i)
\label{eq:15}
\end{equation}
Here, $n_j^v$ represents the number of samples in the $j$-th cluster $C_j$ in the $v$-th view, and $\hat{n}_j$ represents the number of samples in the $j$-th cluster $C_j$ in the fused view.

After some algebraic operations and using the cluster indicator matrix $\mathbf{H}   \in \mathbb{R}^{N\times C}$, $L_{kernel k-means}$ can be expressed as:
\begin{equation}
\begin{split}
L_{kernel k-means}=trace(\hat{\mathbf{K}}(\mathbf{I}-\mathbf{H}\mathbf{H}^T)) \\ + \frac{1}{V}\sum_{v = 1}^{V}trace(\mathbf{K}^v(\mathbf{I}-\mathbf{H}\mathbf{H}^T))
\label{eq:16}
\end{split}
\end{equation}
where $trace(.)$ represents the trace of a matrix, and $\hat{\mathbf{K}}$ and $\mathbf{K}^v$ are the square kernel matrices of the fused view and the original views respectively, given by the following equations:
\begin{equation}
\hat{K}_{ij}=\exp\left(-\frac{1}{\sigma^2}\|\hat{x}_i-\hat{x}_j\|^2\right)
\label{eq:17}
\end{equation}
\begin{equation}
K_{ij}^v=\exp\left(-\frac{1}{\sigma^2}\|x_i^v-x_j^v\|^2\right)
\label{eq:18}
\end{equation}
Given that $\mathbf{H}$ corresponds to the GCN architecture's output, the above loss can be referred to as the multi-view deep kernel k-means loss. Utilizing the kernel k-means objective function helps us exploit the non-linear characteristics generated by the $\mathbf{K}$ matrix to detect linearly separable clusters within the induced space of the individual views and the unified view.

\subsubsection{Spectral Clustering Loss}
The spectral clustering loss enforces the smoothness property of node representations (matrix $\mathbf{H}$ rows) on the graph structure. The loss is expressed as:
\begin{equation}
L_{spectral}=trace(\mathbf{H}^T\mathbf{L}_{f}\mathbf{H})
\label{eq:19}
\end{equation}
where $\mathbf{L}_{f}$ is the Laplacian matrix of the fused graph matrix $\mathbf{A}_{f}$.

\subsubsection{Similarity Matrix Alignment Loss (SMAL)}
Each view in the multi-view dataset is a different representation of the same object. Therefore, the graph structures of each view should be roughly the same. At the same time, to constrain the feature representations of each view to be as aligned as possible in the latent similarity matrix space, we propose the similarity matrix alignment loss:
\begin{equation}
L_{s-alignment}=\sum_{v = 1}^{V}(\|\mathbf{H}\mathbf{H}^T-\mathbf{F}^v(\mathbf{F}^v)^T\|^2+\|\sigma(\mathbf{F}_{f}\mathbf{F}_{f}^T)-\mathbf{F}^v(\mathbf{F}^v)^T\|^2)
\label{eq:20}
\end{equation}
where $\mathbf{S}^v$ is the similarity matrix of the feature matrix $\mathbf{F}^v$ after mapping each view, represented as:
\begin{equation}
\mathbf{S}^v=\mathbf{F}^v(\mathbf{F}^v)^T
\label{eq:21}
\end{equation}

\subsubsection{Feature Representation Alignment Loss (FRAL)}
To ensure that the transformed feature representations retain the original feature information of each view as much as possible, we align the original features with the transformed features through the feature representation alignment loss:
\begin{equation}
L_{f-alignment}=\sum_{v = 1}^{V}\|\mathbf{X}^v(\mathbf{X}^v)^T-\mathbf{S}^v\|^2
\label{eq:22}
\end{equation}

\subsubsection{Overall Loss Function}
Overall, we propose a novel multi-view fusion consensus graph convolutional network for multi-view clustering tasks. The training phase involves joint optimization of the multi-view fusion module, unified graph structure adapter, and GCN feature extraction network according to the following objective function:
\begin{equation}
\begin{split}
L=L_a + \beta \,  L_{kernel k-means}+\lambda_1 \, L_{spectral}+\lambda_2 \,  L_{s-alignment} \\ +\lambda_3 \,  L_{f-alignment}
\label{eq:23}
\end{split}
\end{equation}
where $L_a$ is the Graph Autoencoder loss, which can make the graph $\mathbf{H\, H}^T$ we reconstruct as close as possible to the fused graph $\mathbf{A}_{f}$, and it is represented as follows: 
\begin{equation}
L_a=\|\mathbf{A}_f-\mathbf{H}\mathbf{H}^T\|^2
\label{eq:24}
\end{equation}
In the overall loss function, the hyperparameters $\beta$, $\lambda_1$, $\lambda_2$, and $\lambda_3$ are trade-off parameters.

\subsubsection{Final Learning and Clustering}
The model parameters $\mathbf{U}^v
$, $\mathbf{W}^1$, $\mathbf{W}^2$, and $\mathbf{W}^3$ of MCFCN can be trained by minimizing the overall loss function using the gradient descent method.Following model training, we are able to derive the hidden node representations $\mathbf{H}^1$, $\mathbf{H}^2$, $\mathbf{H}^3$, and then get the orthogonalized output $\mathbf{H}$.

Via training, the hidden node representations $\mathbf{H}^1$ and $\mathbf{H}^2$ implicitly learn the global cluster structure and the orthogonalized output $\mathbf{H}$. In addition, each representation contains information pertaining to a specific neighborhood set.

By integrating \(\mathbf{H}^1\), \(\mathbf{H}^2\), and \(\mathbf{H}\), we construct the final concatenated representation \(\mathbf{H}^F\) to exploit different neighborhood information and global cluster structure knowledge, denoted as:

\begin{equation}
\mathbf{H}^F = [\mathbf{H}^1,\mathbf{H}^2,\mathbf{H}]
\label{eq:25}
\end{equation}
By implementing this strategy, the multi-stage features acquired from the GCN architecture are utilized to enhance the informativeness of the final node representation for clustering. The node partition is then derived by executing the k-means algorithm on \(\mathbf{H}^F\). Algorithm \ref{alg:algorithm1} illustrates the key steps of the MCFCN approach.

\begin{algorithm}[t]
\caption{The Algorithm of MCFCN}
\label{alg:algorithm1}
\KwIn{Multi-view data $\{\mathbf{X}^v\}_{v = 1}^{V}$, number of clusters $C$, number of highest similar samples $k$, dimensions of the linear layers $h_1$ and $h_2$, trade-off parameters $\beta$, $\lambda_1$, $\lambda_2$, $\lambda_3$.}

\KwOut{Clustering result $\mathbf{y}$.}  
\begin{enumerate}
\item Calculate the kernel matrix $\mathbf{K}^v$ using Equation \eqref{eq:18}.
\item While {$\text{epoch}\leq T$}

\item Calculate the unified feature representation $\mathbf{F}_{f}$ through forward propagation using Equations \eqref{eq:1} and \eqref{eq:2}.

\item Calculate the fused adjacency matrix $\mathbf{A}_{f}$ using Equations \eqref{eq:3}, \eqref{eq:4}, and \eqref{eq:5}.

\item Calculate the kernel matrix $\hat{\mathbf{K}}$ using Equation \eqref{eq:17}.

\item Calculate the network outputs $\mathbf{H}^1$, $\mathbf{H}^2$, $\mathbf{H}^3$, and $\mathbf{H}$ through forward propagation using Equations \eqref{eq:6}, \eqref{eq:7}, \eqref{eq:11}, and \eqref{eq:12}.

\item Update the network parameters using Equation \eqref{eq:23}.

\item end while

\item Calculate $\mathbf{H}^F$ using Equation \eqref{eq:25}.

\item Obtain the final clustering result through K-Means on $\mathbf{H}^F$.

\item Return $\mathbf{y}$.

\end{enumerate}
\end{algorithm}

\section{Experimental Setup}
\label{sec:setup}
In this section, we provide a detailed introduction to the datasets, evaluation metrics, and experimental settings used in the experiments. Specifically, we first introduce the multi - view datasets and the comparative methods used in the experiments in Section 4.1. In Section 4.2, we present the evaluation metrics used in this experiment and elaborate on the relevant settings during the training process.

In this part, we provide a detailed introduction to the datasets, algorithms, and other relevant information used in the comparative experiments.

\begin{table*}[h!]
\caption{Statistical specifications of the multi-view datasets.}
\label{tab:multi_view_datasets}
\centering

\begin{tabular}{lrrrc}
\hline
Datasets & Views & Samples & Clusters & Dimensionalities \\
\hline
3Sources & 3 & 169 & 6 & [3560, 3631, 3068] \\
BBCSport & 2 & 544 & 5 & [3183, 3203] \\
Mfeat & 6 & 2000 & 10 & [216, 76, 64, 6, 240, 47] \\
100Leaves & 3 & 1600 & 100 & [64, 64, 64] \\
AWA & 6 & 4000 & 50 & [2688, 2000, 252, 2000, 2000, 2000] \\
NUS & 6 & 2400 & 12 & [64, 144, 73, 128, 225, 500] \\
Caltech101-7 & 6 & 1474 & 7 & [48, 40, 254, 1984, 512, 928] \\
Caltech101-20 & 6 & 2386 & 20 & [48, 40, 254, 1984, 512, 928] \\
\hline
\end{tabular}
\end{table*}

\subsection{Datasets}
For the validation of our model's effectiveness and robustness, we perform experiments on eight widely-adopted benchmark datasets for multi-view clustering (an overview is provided in Table \ref{tab:multi_view_datasets}).

\begin{itemize}
\item 3Sources\footnote{http://mlg.ucd.ie/datasets/3sources.html} is a multi-view dataset with 3 views. It comprises 169 samples and is intended for clustering into 6 clusters. The dimensionalities of its different views are 3560, 3631, and 3068 respectively. This dataset is commonly utilized in multi-view learning research to assess algorithms' capabilities in handling multi-dimensional feature integration and clustering tasks. 

\item BBCSport\footnote{http://mlg.ucd.ie/datasets/segment.html} is a two-view dataset comprising 544 documents derived from five topics: cricket, football, rugby, tennis, and athletics. It is frequently employed in multi-view text analysis and document clustering studies. Researchers use it to evaluate how algorithms process and categorize text data using features from two distinct perspectives. 

\item Mfeat\footnote{http://archive.ics.uci.edu/ml/datasets/Multiple+Features} is a dataset with 6 views, containing 2000 samples that are meant to be grouped into 10 clusters. Its feature dimensionalities are 216, 76, 64, 6, 240, and 47. It serves as a common benchmark in fields like pattern recognition and machine learning algorithm evaluation, helping to gauge how well algorithms deal with multi-source and heterogeneous feature data during clustering. 

\item 100Leaves\footnote{https://archive.ics.uci.edu/ml/datasets/One-hundred+plant+species+
leaves+data+set} is a dataset with 3 views. It has 1600 samples and is designed to be clustered into 100 clusters, with each view having a feature dimensionality of 64. It is mainly applied in image-related research, such as image recognition and the evaluation of clustering algorithms that rely on multi-view features, especially when analyzing and classifying leaf-related images. 

\item AWA \cite{fu2015transductive} is an animal image dataset comprising six distinct views. Initially, it included 30475 images across 50 animal categories, with each image having six feature representations. For experimental purposes, 80 images were randomly selected from each category, resulting in a dataset of 4000 images. It is widely used in computer vision research areas such as image classification and multi-view image feature extraction and clustering. This dataset helps in evaluating how algorithms distinguish and group images of various animal categories based on multiple feature views. 

\item NUS \cite{wang2015deep} is a multi-view dataset with 6 views. It contains 2400 samples and is targeted for clustering into 12 clusters. The feature dimensionalities are 64, 144, 73, 128, 225, and 500. It is often used in multimedia data processing and multi-view data mining research. Scientists leverage it to test algorithms' performance when handling multimedia data with diverse feature types. 

\item Caltech101-7 \cite{fei2004learning} is a dataset with 6 views. It has 1474 samples and is intended for clustering into 7 clusters. Its feature dimensionalities are 48, 40, 254, 1984, 512, and 928. It is mainly applied in image-based research, such as image classification and object recognition tasks that utilize multi-view features. 

\item Caltech101-20 \cite{fei2004learning} is a dataset with 6 views, featuring 2386 samples that are meant to be clustered into 20 clusters. The feature dimensionalities are the same as those of Caltech101-7, i.e., 48, 40, 254, 1984, 512, and 928. It is used in image analysis and computer vision algorithm evaluation.
\end{itemize}

\subsection{Comparison Methods}
We validate the effectiveness of the proposed MCFCN by comparing it with the following eight cutting-edge methods.
\begin{itemize}
\item SiMVC  (Trosten et al. 2021) \cite{trosten2021reconsidering} is a deep multi-view clustering method that utilizes autoencoder networks for acquiring view-specific representations, followed by merging them into a unified final representation.
\item  CoMVC (Trosten et al. 2021) \cite{trosten2021reconsidering} An extension of SiMVC is achieved through the introduction of a selective contrastive learning mechanism for formulating the consensus representation. Herein, positive sample pairs are defined as those samples with the same pseudo-label allocations.
\item CDIMC (Wen et al. 2020) \cite{wen2020CDIMC} employs a cognitive-inspired self-paced K-means clustering component that identifies high-confidence samples, thereby effectively reducing the impact of outliers.
\item DEMVC (XU et al. 2020) \cite{xu2021deep} Through the deep embedding mechanism and view fusion strategy, the feature representations of multiple views are organically integrated to form a unified multi-view joint representation, effectively integrating information from different views and enabling accurate clustering.
\item MFLVC (Xu et al. 2022) \cite{xu2022multi} applies instance-level and cluster-level contrastive objectives concurrently, enhancing feature extraction and clustering performance in an end-to-end fashion.
\item GCFAgg (Yan et al. 2023) \cite{yan2023gcfagg} integrates a contrastive learning-driven module to learn the global sample-wise dependencies, coercing view-specific representations of highly related instances to converge toward similar feature spaces.
\item SURER (Wang et al. 2024) \cite{wang2024surer} employs a graph structure learning module and a heterogeneous unified graph neural network to optimize the graph topology, then leverages complementary information across views to learn the consensus representation.
\item DFP-GNN (Xiao et al. 2023) \cite{xiao2023dual} a novel dual-fusion combined with dual-information-propagation mechanism is developed, aiming to comprehensively exploit the complementary consistency within multi-view datasets.
\end{itemize}

\subsection{Evaluation Metrics and Training}
We employ four widely-used metrics to evaluate the clustering performance of different methods on the same dataset, including clustering accuracy (ACC) \cite{cai2010locally}, normalized mutual information (NMI) \cite{kuipers1975frame}, adjusted Rand index (ARI) \cite{hopkins2018big}, and F1-score (F1) \cite{amigo2009comparison}. Their mathematical definitions are as follows.
\begin{equation}
\mathrm{ACC}(\mathbf{y}_{\mathrm{true}}, \mathbf{y}_{\mathrm{pred}}) = \frac{\sum_{i = 1}^{N} \mathbf{1}[\mathbf{y}_{\mathrm{true}}(i) = T(\mathbf{y}_{\mathrm{pred}}(i))]}{N}
\end{equation}
\begin{equation}
NMI(\mathbf{y}_{\mathrm{true}}, \mathbf{y}_{\mathrm{pred}}) = \frac{MI(\mathbf{y}_{\mathrm{true}}, \mathbf{y}_{\mathrm{pred}})}{\frac{1}{2}[En(\mathbf{y}_{\mathrm{true}}) + En(\mathbf{y}_{\mathrm{pred}})]}
\end{equation}
\(\mathbf{y}_{\mathrm{true}}\) represents the vector of true labels, \(\mathbf{y}_{\mathrm{pred}}\) represents the vector of clustering indices predicted by the model, and $N$ represents the total number of samples, $MI$ represents the mutual information function, and $En$ represents entropy. The Hungarian algorithm is adopted to find the optimal mapping $T$ between the true labels and the predicted labels.
\begin{equation}
ARI = \frac{RI-\mathrm{expected}(RI)}{\mathrm{max}(RI)-\mathrm{expected}(RI)}
\end{equation}
\begin{equation}
RI = \frac{n_1 + n_2}{n_1 + n_2 + n_3 + n_4}
\end{equation}

Let \( S = \{o_1, o_2, \ldots, o_n\} \) be a set with two partitions \( X = \{X_1, X_2, \ldots, X_r\} \) and \( Y = \{Y_1, Y_2, \ldots, Y_s\} \). The terms \( n_1 \), \( n_2 \), \( n_3 \), and \( n_4 \) are defined as follows:

\begin{itemize}
   \item \( n_1 \): The number of element pairs in \( S \) that are in the same cluster  in both partition \( X \) and partition \( Y \).
   \item \( n_2 \): The number of element pairs in \( S \) that are in different clusters in both partition \( X \) and partition \( Y \).
  \item \( n_3 \): The number of element pairs in \( S \) that are in the same cluster in partition \( X \) but in different clusters in partition \( Y \).
    \item \( n_4 \): The number of element pairs in \( S \) that are in different clusters in partition \( X \) but in the same cluster in partition \( Y \).
\end{itemize}

In our experiments, we initialize the weights of the linear layers—including the two GCN layers and the final linear layer—randomly. The output dimension of the multi-view feature fusion module, i.e., the first linear layer, is set to 256. For the GCN Feature Extraction Network, the hidden layer dimensions are set as $h_1 = h_2 = 16$. During the computation of  \(\mathbf{A}_f\), we set $k = 10$. The MCFCN model is trained using the Adam optimizer with a learning rate of 0.001, and all experiments are implemented using PyTorch.

To ensure a fair comparison, all baseline methods are reproduced using the official source codes provided by the authors and configured with the optimal parameters recommended in their respective publications.

\begin{table}
\caption{The clustering performance of all methods on eight datasets. For each dataset, the best results are shown in bold and the second-best results are underlined.
}
\label{tab:result_1}
\centering
\resizebox{0.44\textwidth}{!}{
\begin{tabular}{clcccc}
\hline
Datasets & Methods & ACC & NMI & ARI & F1 \\
\hline
\multirow{9}{*}{3Sources}
& SiMVC \cite{trosten2021reconsidering} & 0.2781 & 0.0757 & 0.2339 & 0.2675\\
& CoMVC \cite{trosten2021reconsidering} & 0.2722 & 0.0428 & 0.1968 & 0.2712\\
& CDIMC \cite{wen2020CDIMC} & 0.4734 & 0.4688 & 0.3021 & 0.2282\\
& DEMVC \cite{xu2021deep} & 0.5444 & 0.3972 & 0.2887 & 0.4671\\
& MFLVC \cite{xu2022multi} & 0.5325 & 0.5298 & 0.3465 & 0.5366\\
& GCFAgg \cite{yan2023gcfagg} & 0.5207 & 0.3644 & 0.2282 & 0.4256\\
& SURER \cite{wang2024surer} & \underline{0.7337} & \underline{0.6091} & \underline{0.6331} & \underline{0.6729}\\
& DFP-GNN \cite{xiao2023dual} & 0.645 & 0.6038 & 0.5954 & 0.6403\\
& MCFCN (Ours) & \textbf{0.8402} & \textbf{0.7406} & \textbf{0.6797} & \textbf{0.8342}\\
\bottomrule
\multirow{9}{*}{BBCSport}
& SiMVC \cite{trosten2021reconsidering}	& 0.2959 & 0.0349 & 0.0338 & 0.2481\\
& CoMVC \cite{trosten2021reconsidering} & 0.2886 & 0.0433 & 0.0226 & 0.2496\\
& CDIMC \cite{wen2020CDIMC}	& 0.4099 & 0.2607 & 0.095 & 0.4981\\
& DEMVC \cite{xu2021deep}	& 0.4632 & 0.2760 & 0.1983 & 0.4261\\
& MFLVC \cite{xu2022multi}	& 0.6301 & 0.4361 & 0.3722 & 0.5148\\
& GCFAgg \cite{yan2023gcfagg} & 0.4375 & 0.3427 & 0.2411 & 0.4332\\
& SURER \cite{wang2024surer} & \underline{0.9632} & \underline{0.8911} & \underline{0.9013} & \underline{0.9295}\\
& DFP-GNN \cite{xiao2023dual} & 0.8676 & 0.7577 & 0.7073 & 0.8018\\
& MCFCN (Ours) & \textbf{0.9651} & \textbf{0.8964} & \textbf{0.9108} & \textbf{0.9632}\\
\bottomrule
\multirow{9}{*}{Mfeat}
& SiMVC \cite{trosten2021reconsidering} & 0.8385 & 0.7994 & 0.7188 & 0.7561\\
& CoMVC \cite{trosten2021reconsidering} & 0.815 & 0.8228 & 0.7331 & 0.7897\\
& CDIMC \cite{wen2020CDIMC} & 0.8430 & 0.8913 & 0.8146 & 0.8760\\
& DEMVC \cite{xu2021deep} & 0.2335 & 0.1806 & 0.0814 & 0.2638\\
& MFLVC \cite{xu2022multi} & 0.8575 & 0.8272 & 0.8583 & 0.7767\\
& GCFAgg \cite{yan2023gcfagg} & 0.8155 & 0.7368 & 0.6628 & 0.7095\\
& SURER \cite{wang2024surer} & \textbf{0.9665} & \textbf{0.9348} & \textbf{0.9254} & \underline{0.9385}\\
& DFP-GNN \cite{xiao2023dual} & 0.842 & 0.8339 & 0.7637 & 0.7948\\
& MCFCN (Ours) & \underline{0.9637} & \underline{0.9246} & \underline{0.9202} & \textbf{0.9632}\\
\bottomrule
\multirow{9}{*}{100Leaves}
& SiMVC \cite{trosten2021reconsidering} & 0.5031 & 0.7862 & 0.3965 & 0.5074\\
& CoMVC \cite{trosten2021reconsidering} & 0.5518 & 0.7981 & 0.4293 & 0.5435\\
& CDIMC \cite{wen2020CDIMC} & 0.7744 & 0.9276 & \underline{0.8093} & 0.7090\\
& DEMVC \cite{xu2021deep} & 0.3675 & 0.3822 & 0.3118 & 0.2916\\
& MFLVC \cite{xu2022multi} & 0.7613 & 0.8775 & 0.7485 & 0.6722\\
& GCFAgg \cite{yan2023gcfagg} & 0.2238 & 0.5678 & 0.1182 & 0.1902\\
& SURER \cite{wang2024surer} & \underline{0.8579} & \underline{0.9352} & 0.7788 & \underline{0.8366}\\
& DFP-GNN \cite{xiao2023dual} & 0.5419 & 0.8025 & 0.4424 & 0.529\\
& MCFCN (Ours) & \textbf{0.9656} & \textbf{0.9791} & \textbf{0.9363} & \textbf{0.9652}\\
\hline
\multirow{9}{*}{AWA}
& SiMVC \cite{trosten2021reconsidering} & 0.1072 & 0.1806 & 0.0314 & 0.0735\\
& CoMVC \cite{trosten2021reconsidering} & 0.1077 & 0.1775 & 0.0324 & 0.0744\\
& CDIMC \cite{wen2020CDIMC} & 0.0998 & 0.1737 & 0.0133 & \underline{0.1039}\\
& DEMVC \cite{xu2021deep} & 0.0592 & 0.0960 & 0.0153 & 0.0594\\
& MFLVC \cite{xu2022multi} & 0.0688 & 0.1026 & 0.0077 & 0.0538\\
& GCFAgg \cite{yan2023gcfagg} & \underline{0.116} & \underline{0.1987} & \underline{0.0418} & 0.0712\\
& SURER \cite{wang2024surer} & 0.0837 & 0.1248 & 0.0135 & 0.0597\\
& DFP-GNN \cite{xiao2023dual} & 0.0843 & 0.1548 & 0.0208 & 0.0579\\
& MCFCN (Ours) & \textbf{0.1198} & \textbf{0.2055} & \textbf{0.0425} & \textbf{0.1200}\\
\bottomrule
\multirow{9}{*}{NUS}
& SiMVC \cite{trosten2021reconsidering} & 0.2583 & 0.1461 & 0.0816 & 0.1731\\
& CoMVC \cite{trosten2021reconsidering} & 0.2525 & 0.1264 & 0.0688 & 0.1619\\
& CDIMC \cite{wen2020CDIMC} & 0.2421 & 0.1276	& 0.0565 & 0.1764\\
& DEMVC \cite{xu2021deep} & 0.1892 & 0.0829 & 0.0335 & 0.1340\\
& MFLVC \cite{xu2022multi} & \textbf{0.3013} & \textbf{0.1677} & \underline{0.0987} & 0.1799\\
& GCFAgg \cite{yan2023gcfagg} & \underline{0.2821} & 0.1590 & 0.0958 & 0.1779\\
& SURER \cite{wang2024surer} & 0.2633 & \underline{0.1653} & 0.0803 & \underline{0.2082}\\
& DFP-GNN \cite{xiao2023dual} & 0.2121 & 0.1071 & 0.0507 & 0.153\\
& MCFCN (Ours) & \textbf{0.3013} & 0.1554 & \textbf{0.1001} & \textbf{0.2973}\\
\bottomrule
\multirow{9}{*}{Caltech101-7}
& SiMVC \cite{trosten2021reconsidering} & 0.3758 & 0.3859 & 0.2248 & 0.4431\\
& CoMVC \cite{trosten2021reconsidering} & 0.3629 & 0.3721 & 0.2182 & 0.4373\\
& CDIMC \cite{wen2020CDIMC} & 0.5237 & 0.5663 & 0.3925 & 0.5871\\
& DEMVC & 0.6004 & 0.4888 & \underline{0.4650} & \underline{0.6464}\\
& MFLVC \cite{xu2022multi} & 0.5095 & \underline{0.6065} & 0.3980 & 0.6092\\
& GCFAgg \cite{yan2023gcfagg} & 0.4512 & 0.5629 & 0.3336 & 0.5656\\
& SURER \cite{wang2024surer} & \underline{0.6588} & 0.4096 & 0.3475 & 0.6228\\
& DFP-GNN \cite{xiao2023dual} & 0.4322 & 0.583 & 0.3562 & 0.5774\\
& MCFCN (Ours) & \textbf{0.8881} & \textbf{0.7596} & \textbf{0.8461} & \textbf{0.6616}\\
\bottomrule
\multirow{9}{*}{Caltech101-20}
& SiMVC \cite{trosten2021reconsidering} & 0.3168 & 0.4791 & 0.2499 & 0.3691\\
& CoMVC \cite{trosten2021reconsidering} & 0.2979 & 0.4092 & 0.1761 & 0.3233\\
& CDIMC \cite{wen2020CDIMC} & \underline{0.5218} & \underline{0.6557} & \underline{0.4009} & \underline{0.5697}\\
& DEMVC & 0.3734 & 0.2423 & 0.2052 & 0.3462\\
& MFLVC \cite{xu2022multi} & 0.4028 & 0.4734 & 0.3486 & 0.4219\\
& GCFAgg \cite{yan2023gcfagg} & 0.3734 & 0.545 & 0.2689 & 0.4359\\
& SURER \cite{wang2024surer} & 0.4807 & 0.5218 & 0.3621 & 0.5036\\
& DFP-GNN \cite{xiao2023dual} & 0.4677 & 0.6372 & 0.3746 & 0.5349\\
& MCFCN (Ours) & \textbf{0.7402} & \textbf{0.6894} & \textbf{0.7281} & \textbf{0.5099}\\
\hline
\end{tabular}
}
\end{table}

\section{Performance Evaluation }
\label{sec:Performance}
In this section, we evaluate the performance of MCFCN on multi-view clustering tasks through extensive experiments. Specifically, we first introduce and analyze the results of a large number of experiments conducted on eight multi-view datasets in Section \ref{sec:Performance}.1  to verify the effectiveness of the proposed MCFCN. In Section \ref{sec:Performance}.2, we analyze the effectiveness of each component. In Section 4.3, we analyze the hyperparameters of the model.

\subsection{ Performance and Analysis}
The comparative results of eight multi-view clustering methods using four evaluation metrics (ACC, NMI, ARI, F1) on eight benchmark datasets of different scales are shown in Table \ref{tab:result_1}.  From the experimental results, it can be seen that the proposed MCFCN method generally achieves better results compared with other methods. Specifically, we draw the following observations:
\begin{enumerate}
    \renewcommand{\theenumi}{\arabic{enumi}} 
    \item The proposed method demonstrates significant superiority over all compared methods on most datasets and achieves the highest metric scores across eight datasets. This proves the effectiveness of the Multi - view Feature Fusion Module, Unified Graph Structure Adapter, and GCN Feature Extraction Network proposed in our method. They can effectively represent the consistent structure of multi - views and obtain the corresponding feature representations.
    \item We compared the proposed method with four state-of-the-art deep multi-view clustering methods (DEMVC, SiMVC, CoMVC, and MFLVC). DEMVC aligns the label distributions of individual views with a target distribution, which may inadvertently disrupt the original graph structure and degrade feature representation. SiMVC and CoMVC adopt view-wise fusion strategies to obtain a consensus representation, but private information from each view may overshadow discriminative features during the fusion process. In contrast, our proposed MCFCN method enhances cross-view structural consistency, leading to more discriminative clustering results. Experimental results demonstrate that MCFCN outperforms these baselines across all evaluated datasets.
    \item Compared with other graph- and graph neural network-based multi-view clustering methods (e.g., SURER, DFP-GNN, and GCFAgg), our proposed approach consistently achieves the best clustering results across datasets. This demonstrates that the Unified Graph Structure Adapter in MCFCN effectively captures cross-view structural consistencies, thereby enhancing the representational power of GCN features.
\end{enumerate}

\begin{figure*}[h!]
  \centering
  \subfigure[The consensus graph matrix of BBCSport]{\includegraphics[height=1.6in, width=1.75in]{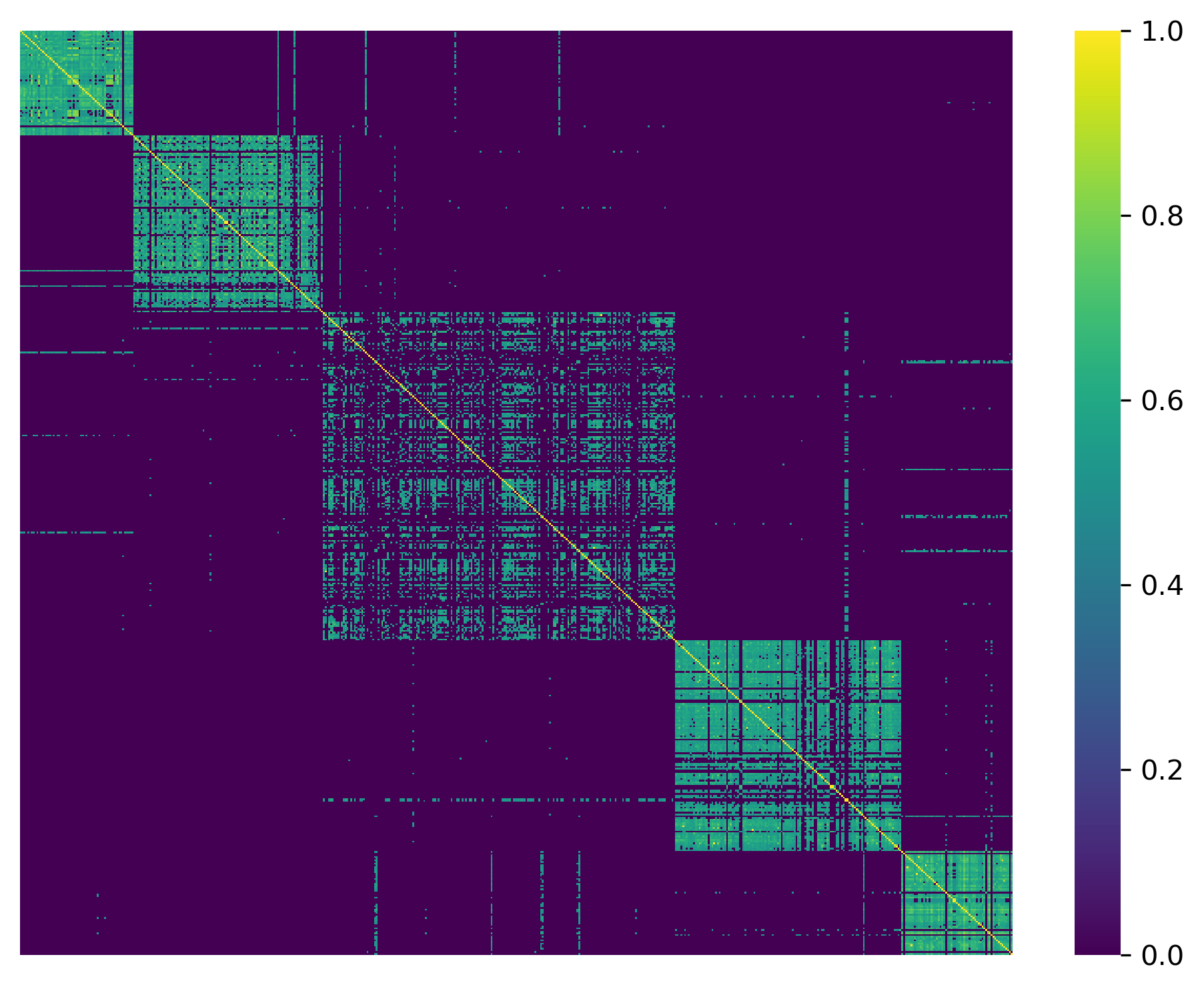}} 
  \subfigure[The consensus graph matrix  of 100Leaves]{\includegraphics[height=1.6in, width=1.75in]{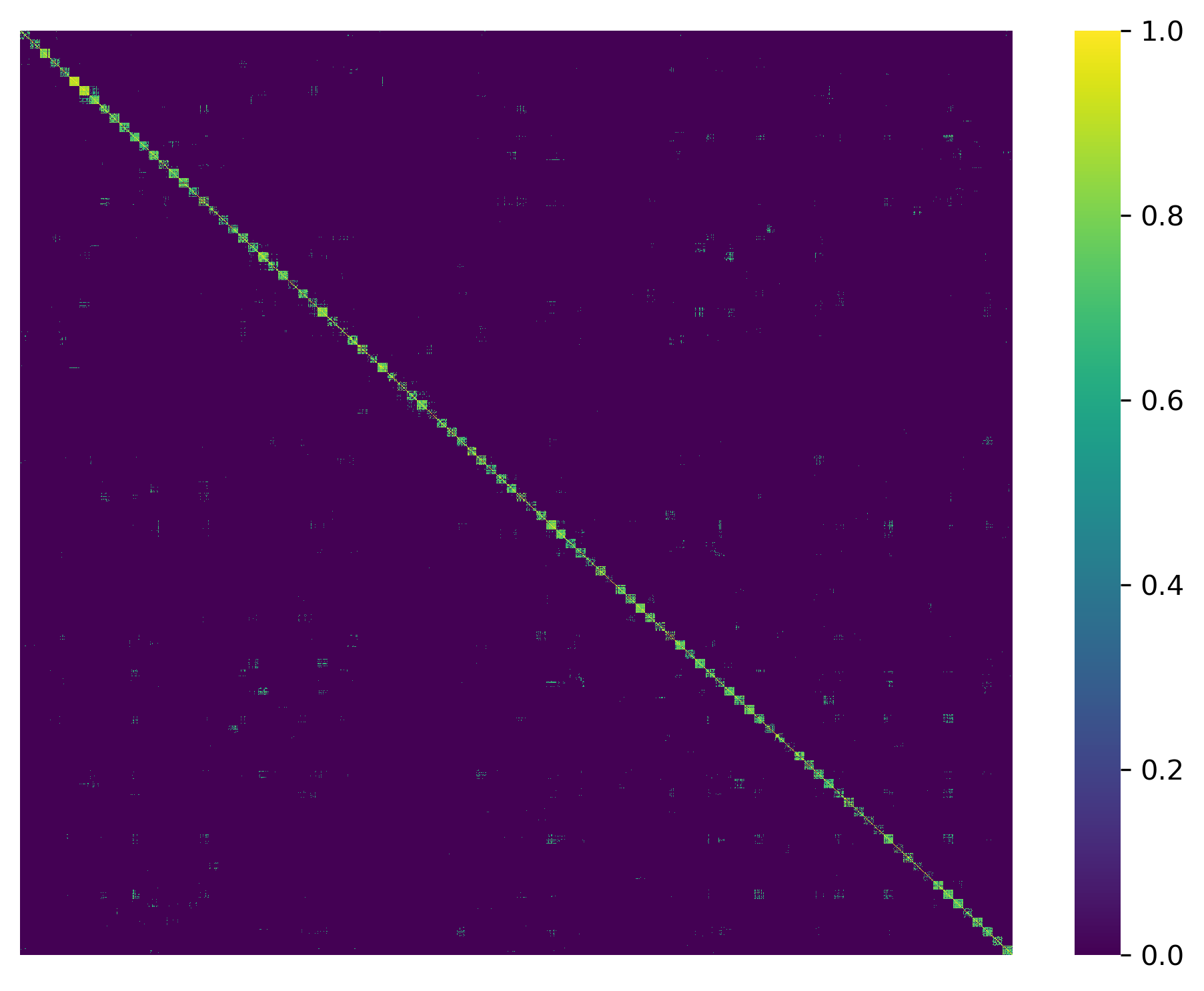}} 
  \subfigure[The consensus graph matrix  of Mfeat]{\includegraphics[height=1.6in, width=1.75in]{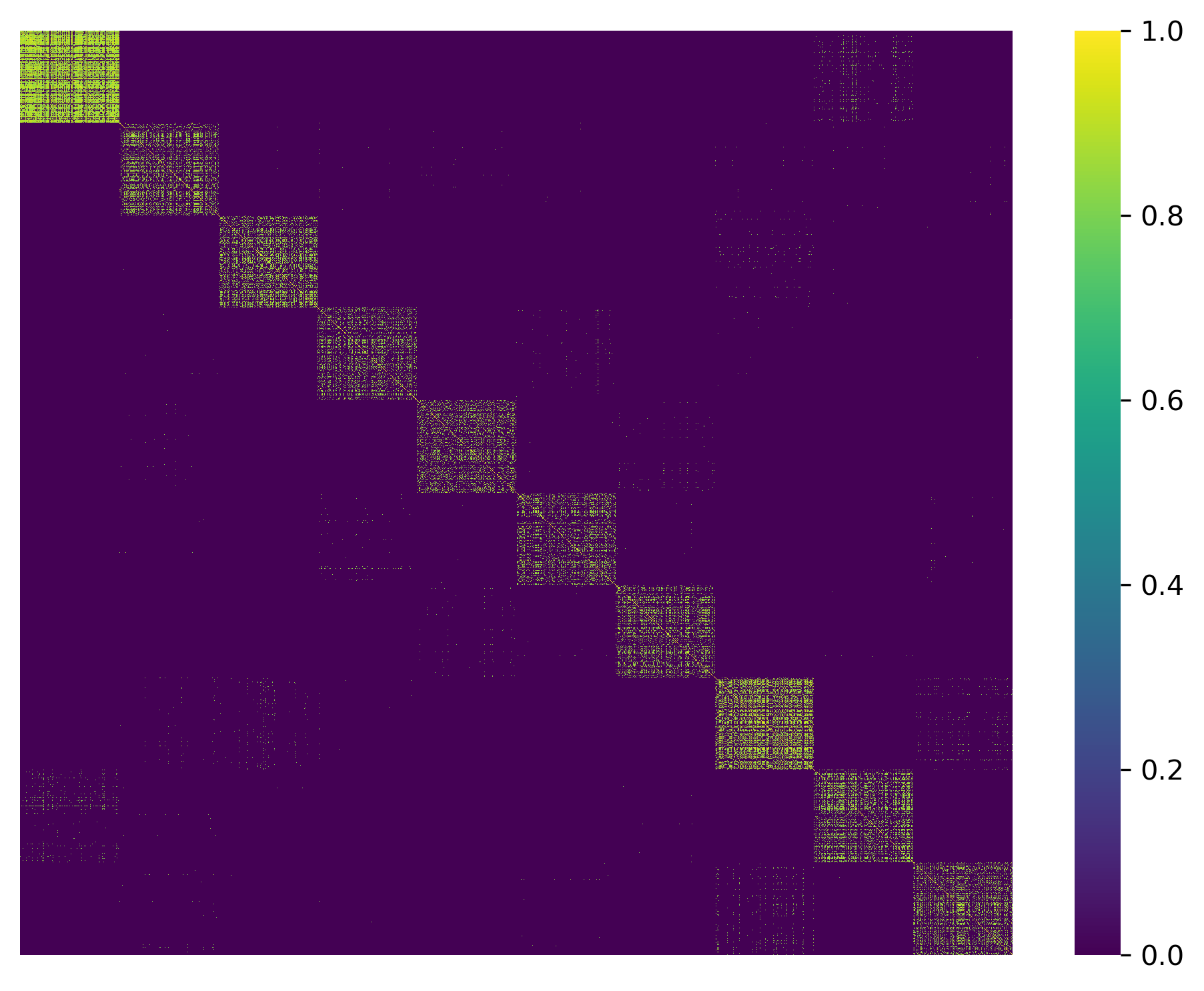}} 
  \subfigure[The consensus graph matrix of Caltech101-7]{\includegraphics[height=1.6in, width=1.75in]{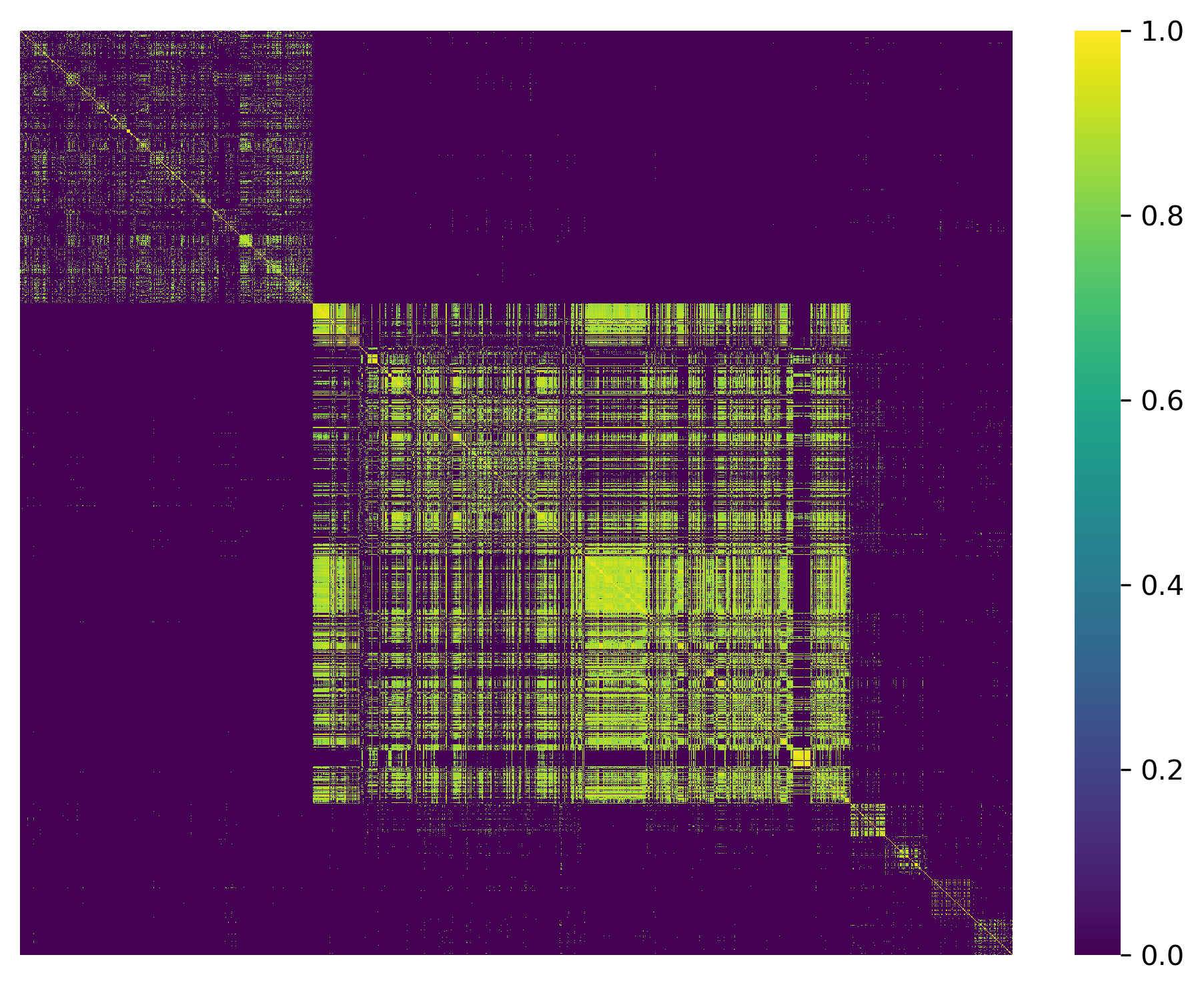}} 
  \caption{Visualization of the consensus graph in the BBCSport, 100Leaves, Mfeat and Caltech101-7.}
  \label{graph_hot}
\end{figure*}

Constructing the consensus graph structure constitutes a core component of our approach and serves as a critical input for the GCN.  To illustrate the quality  of the estimated  consensus graph matrix $\mathbf{A}_{f}$, we conducted visualization. Figure \ref{graph_hot} illustrates the consensus graphs learned by our algorithm on the BBCSport, 100Leaves, Mfeat, and Caltech101-7 datasets. As illustrated in the figure, all four graphs exhibit distinct block structures that roughly align with the number of clusters in the datasets. These findings demonstrate that the proposed algorithm effectively learns the multi-view consensus graph structure from the datasets.

\begin{figure*}[h!]
  \setlength{\fboxsep}{0pt} 
  \setlength{\fboxrule}{0.1pt} 
  \centering
  \subfigure[Raw Features]{\fbox{\includegraphics[height=1.5in, width=1.7in]{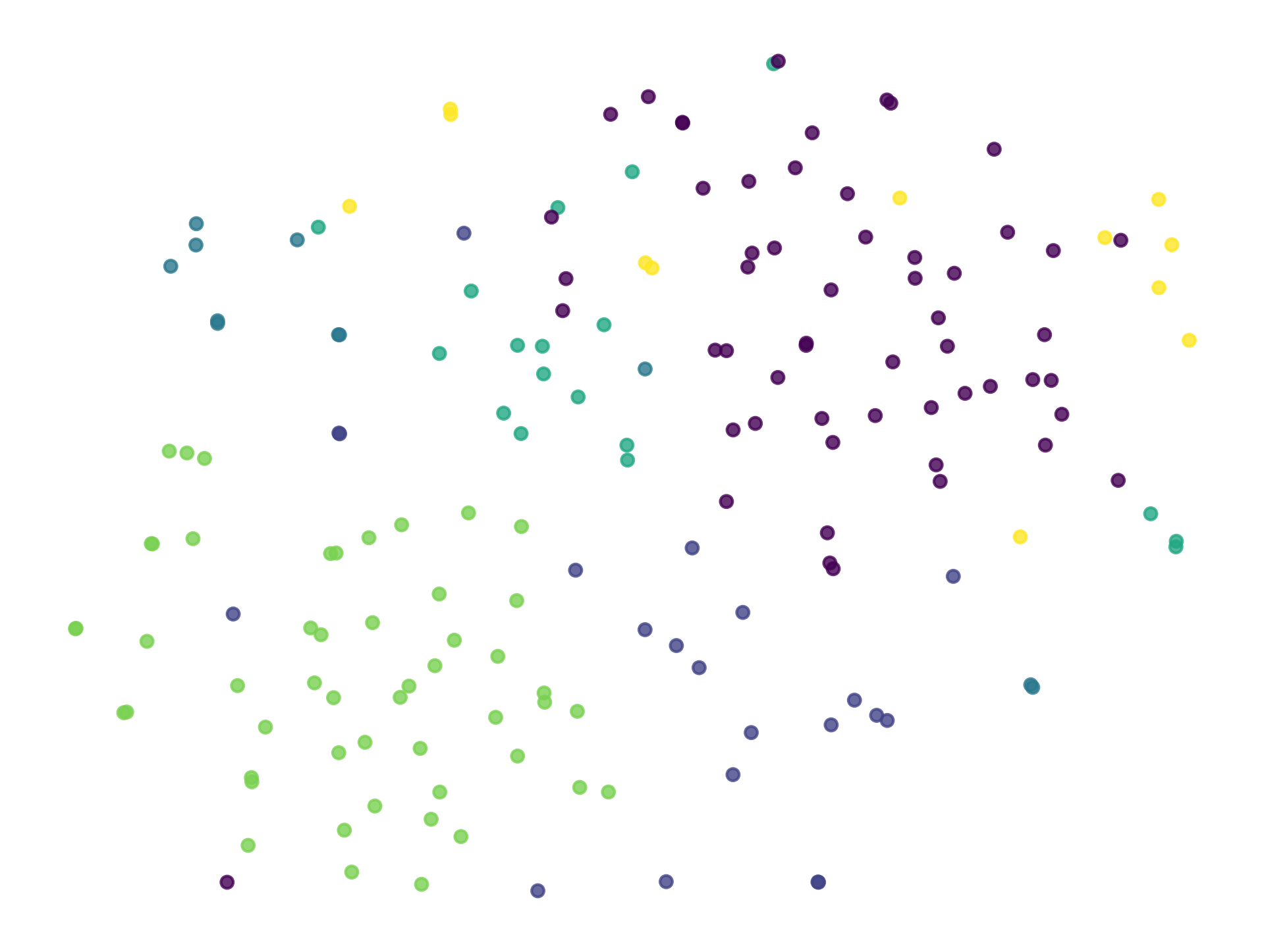}}}
  \subfigure[MFLVC]{\fbox{\includegraphics[height=1.5in, width=1.7in]{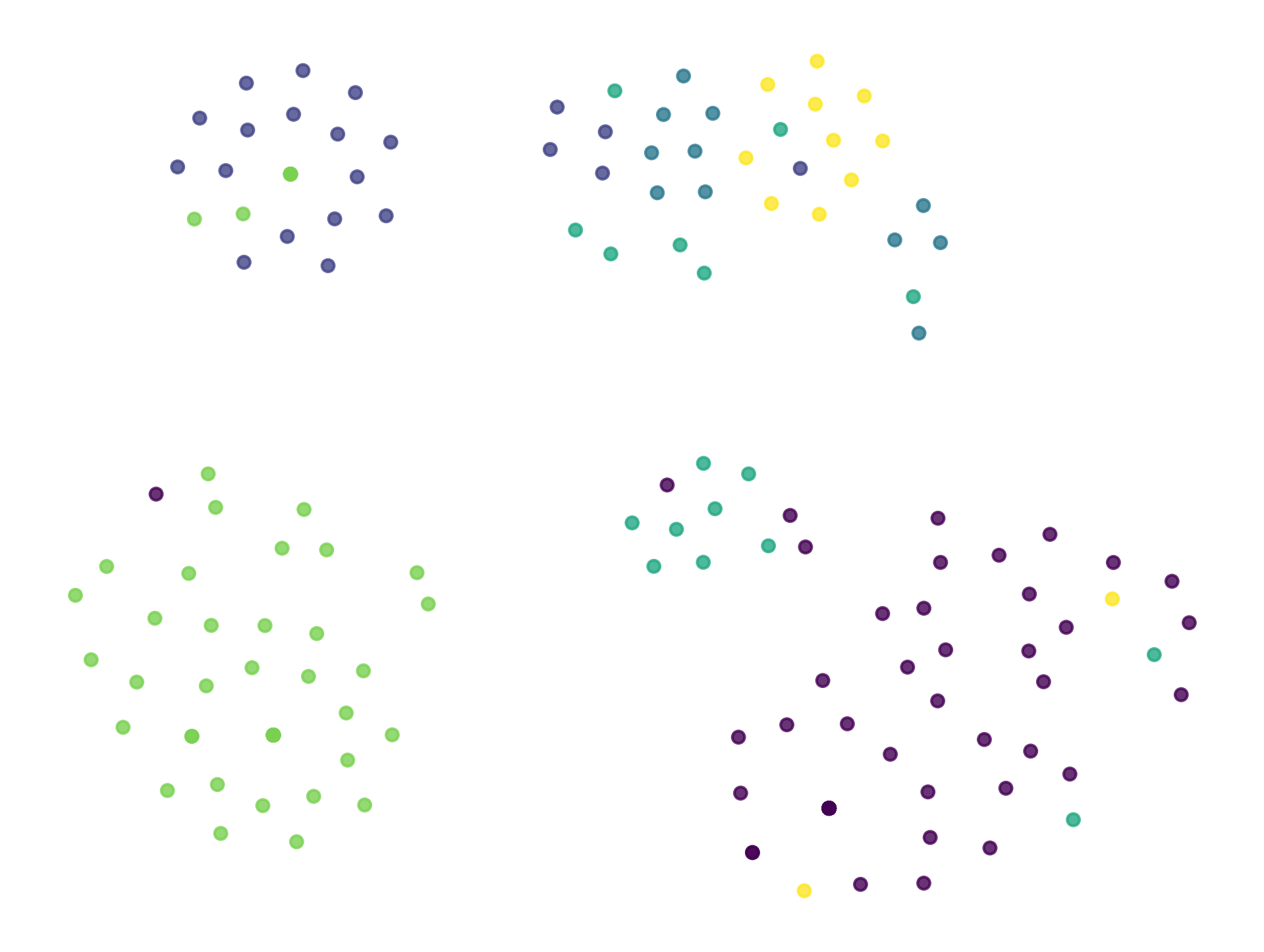}}} 
  \subfigure[SURER]{\fbox{\includegraphics[height=1.5in, width=1.7in]{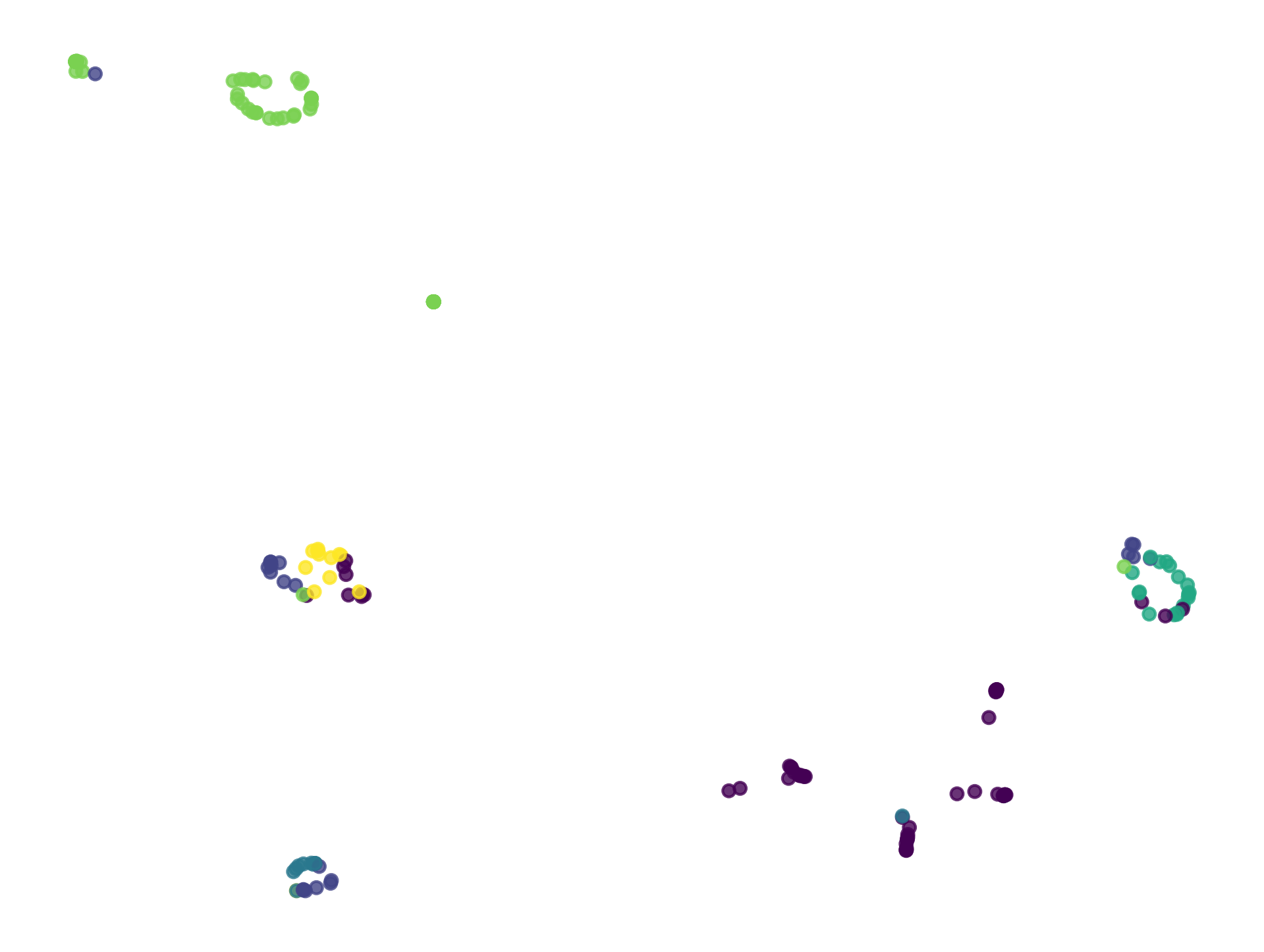}}} 
  \subfigure[Our MCFCN]{\fbox{\includegraphics[height=1.5in, width=1.7in]{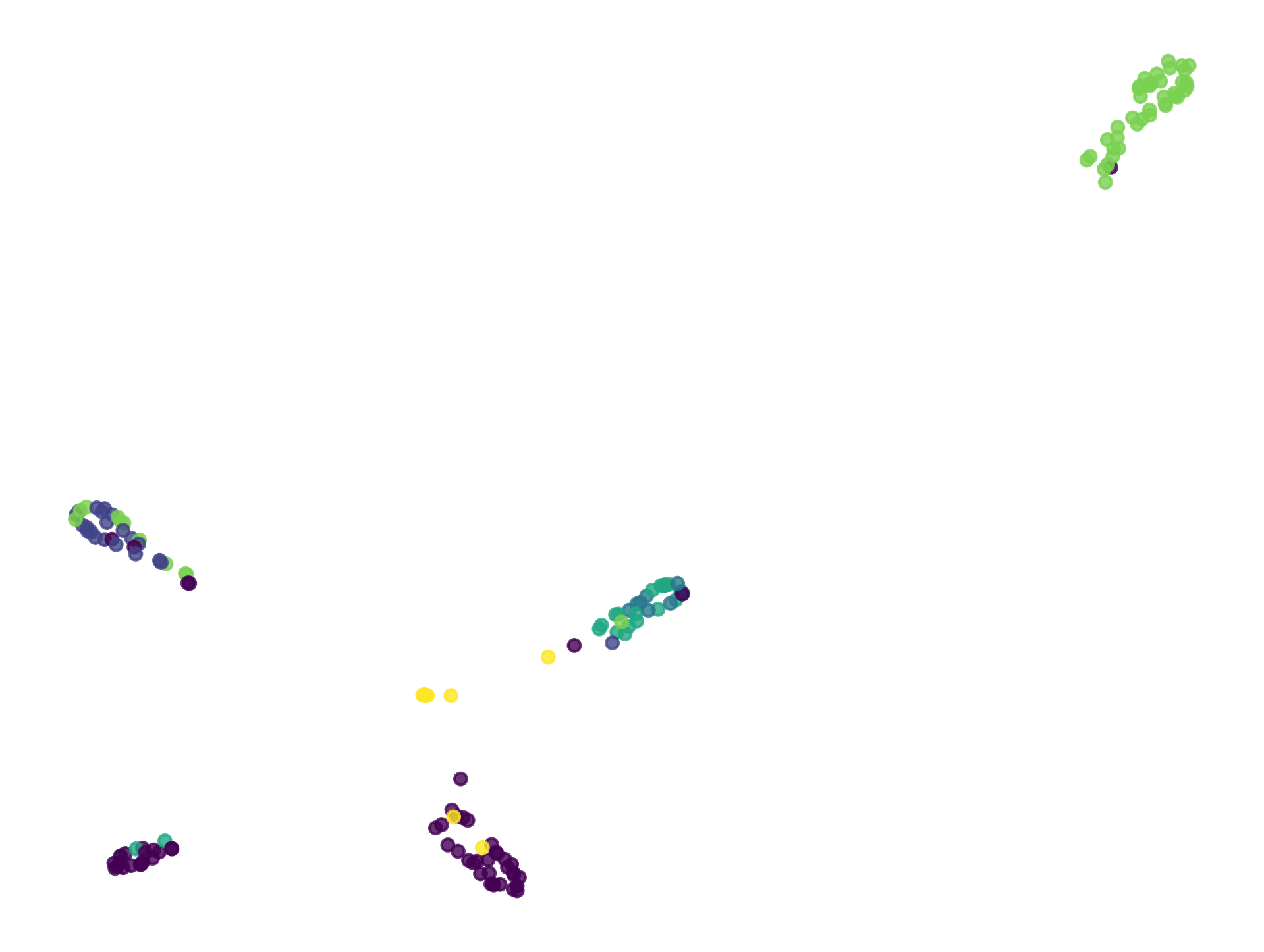}}} 
  \caption{The t-SNE visualization of the raw features and the learned representation by different methods on the 3Sources dataset.}
  \label{3c_tSNE}
\end{figure*}

\begin{figure*}[h!]
  \setlength{\fboxsep}{0pt} 
  \setlength{\fboxrule}{0.1pt} 
  \centering
  \subfigure[Raw Features]{\fbox{\includegraphics[height=1.5in, width=1.7in]{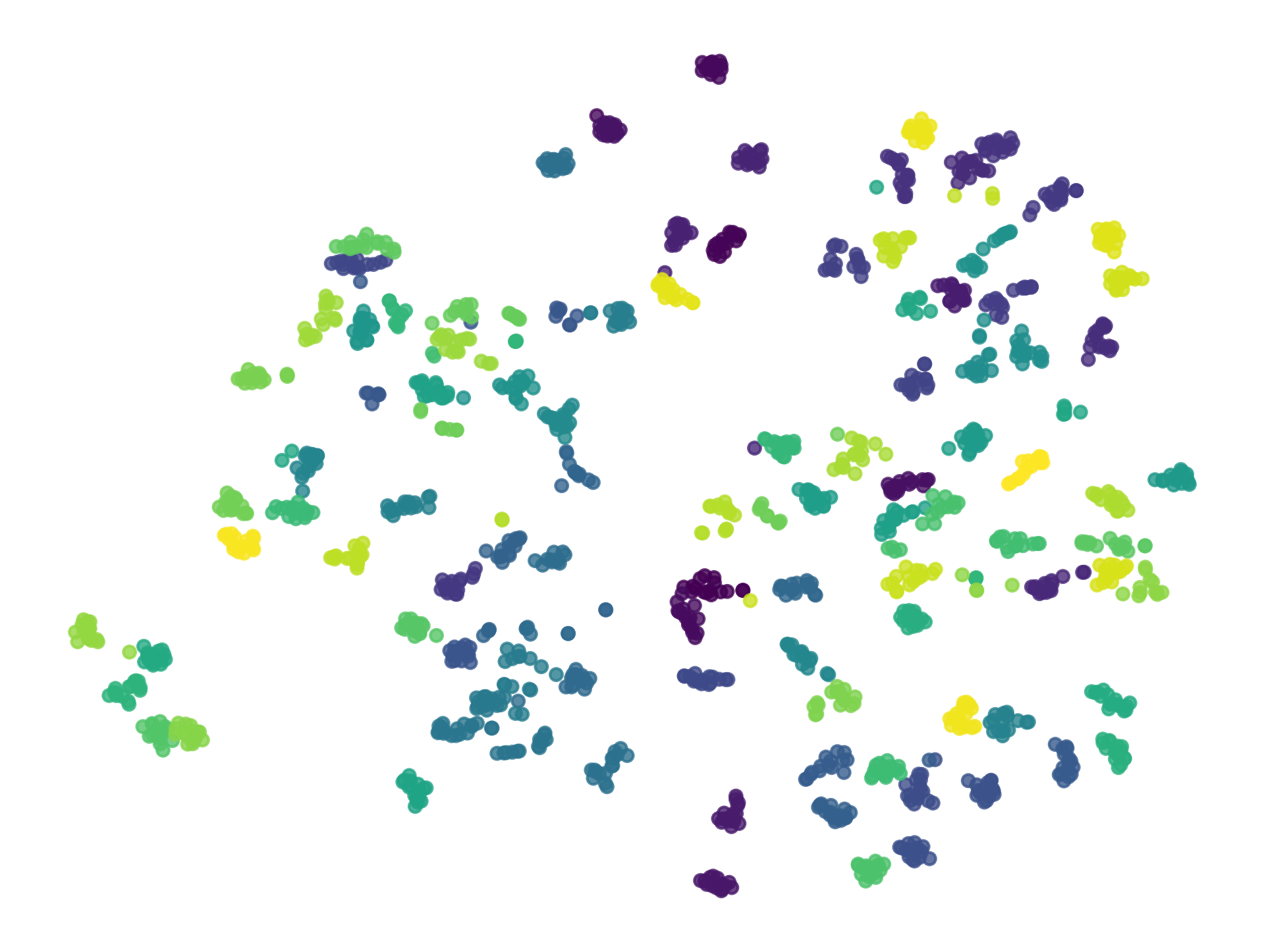}}}
  \subfigure[MFLVC]{\fbox{\includegraphics[height=1.5in, width=1.7in]{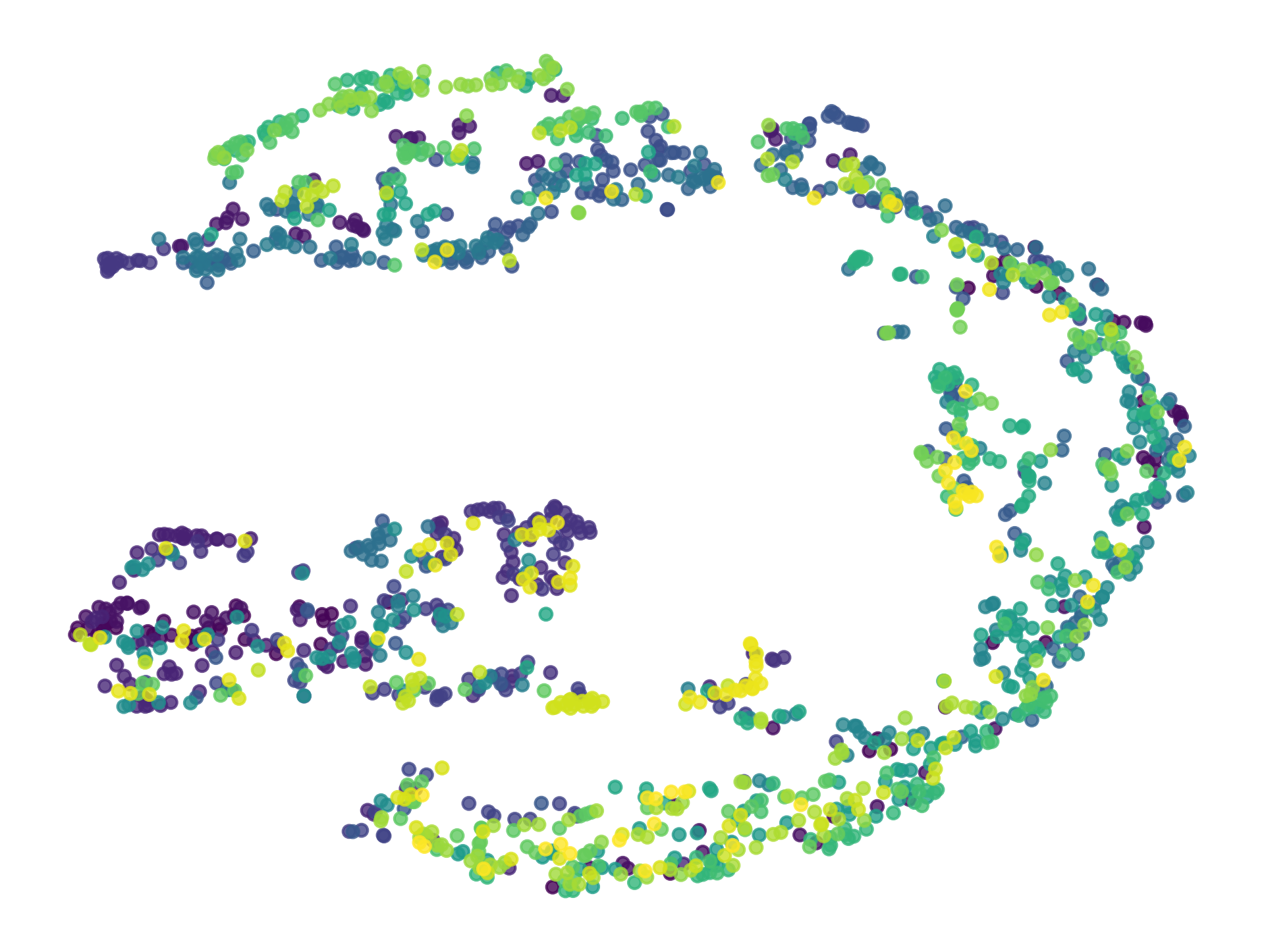}}} 
  \subfigure[SURER]{\fbox{\includegraphics[height=1.5in, width=1.7in]{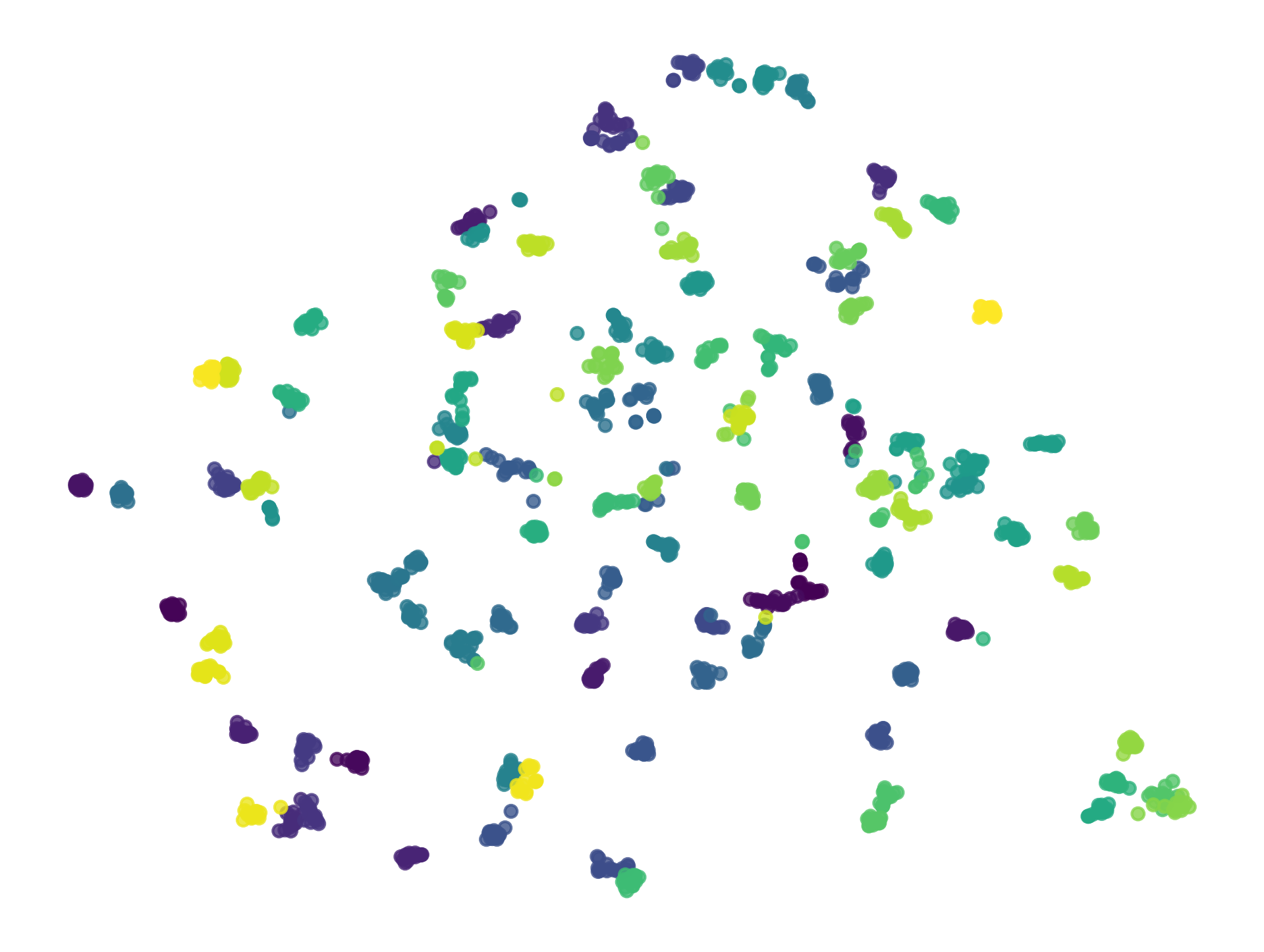}}} 
  \subfigure[Our MCFCN]{\fbox{\includegraphics[height=1.5in, width=1.7in]{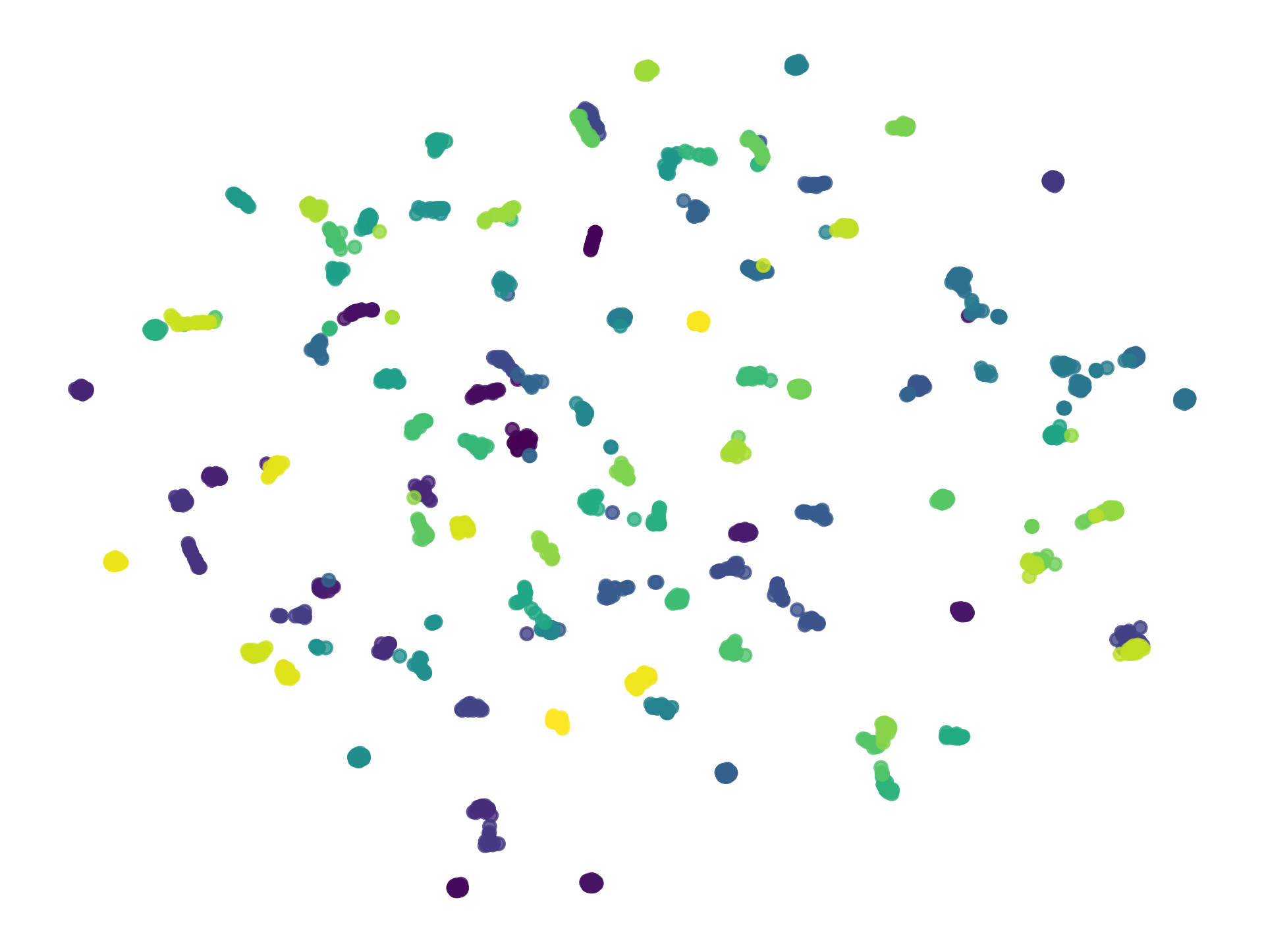}}} 
  \caption{ The t-SNE visualization of the raw features and the learned representation by different methods on the 100Leaves.}
  \label{100_tSNE}
\end{figure*}

\begin{figure*}[h!]
  \setlength{\fboxsep}{0pt} 
  \setlength{\fboxrule}{0.1pt} 
  \centering
  \subfigure[Raw Features]{\fbox{\includegraphics[height=1.5in, width=1.7in]{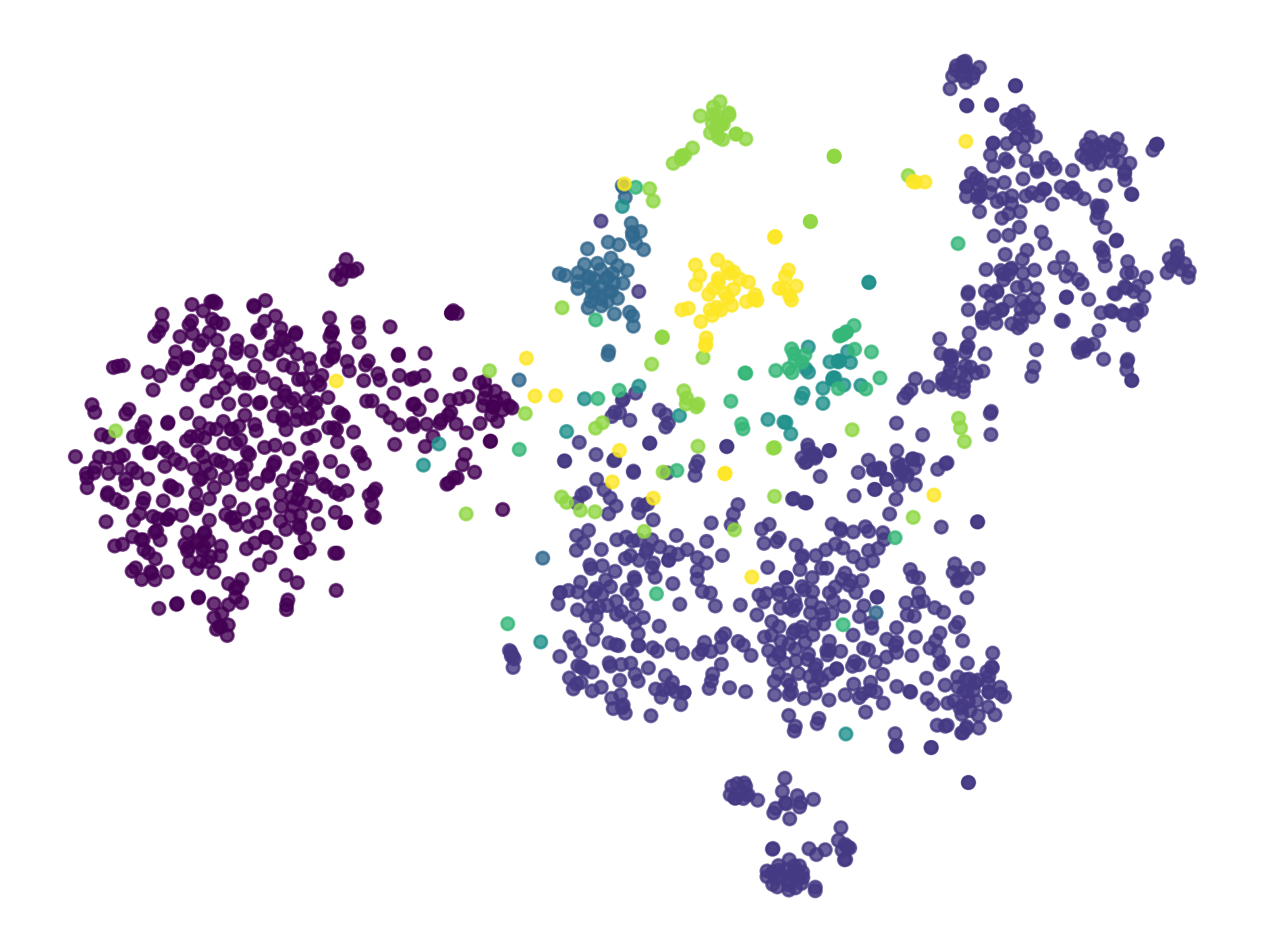}}}
  \subfigure[MFLVC]{\fbox{\includegraphics[height=1.5in, width=1.7in]{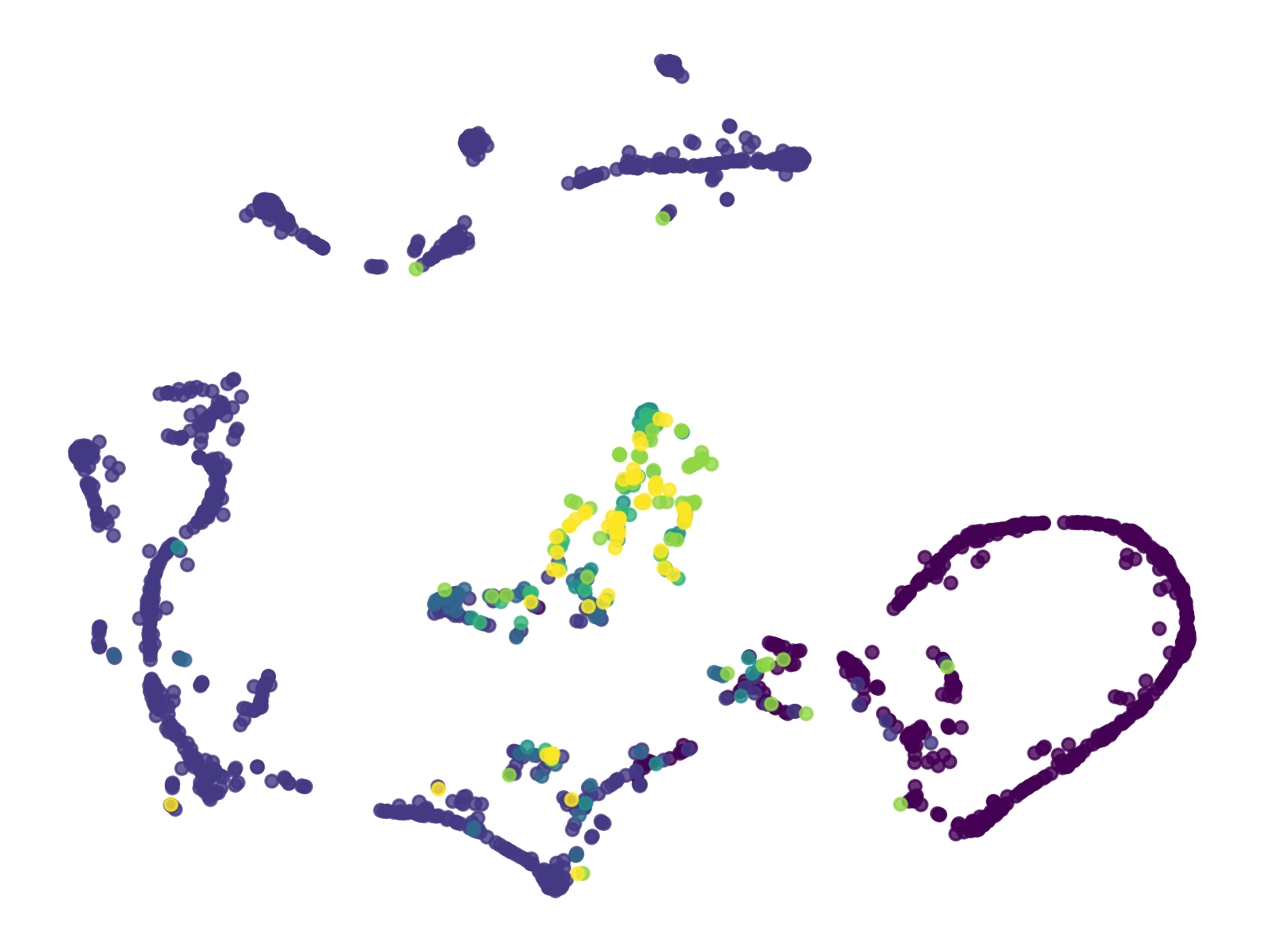}}} 
  \subfigure[SURER]{\fbox{\includegraphics[height=1.5in, width=1.7in]{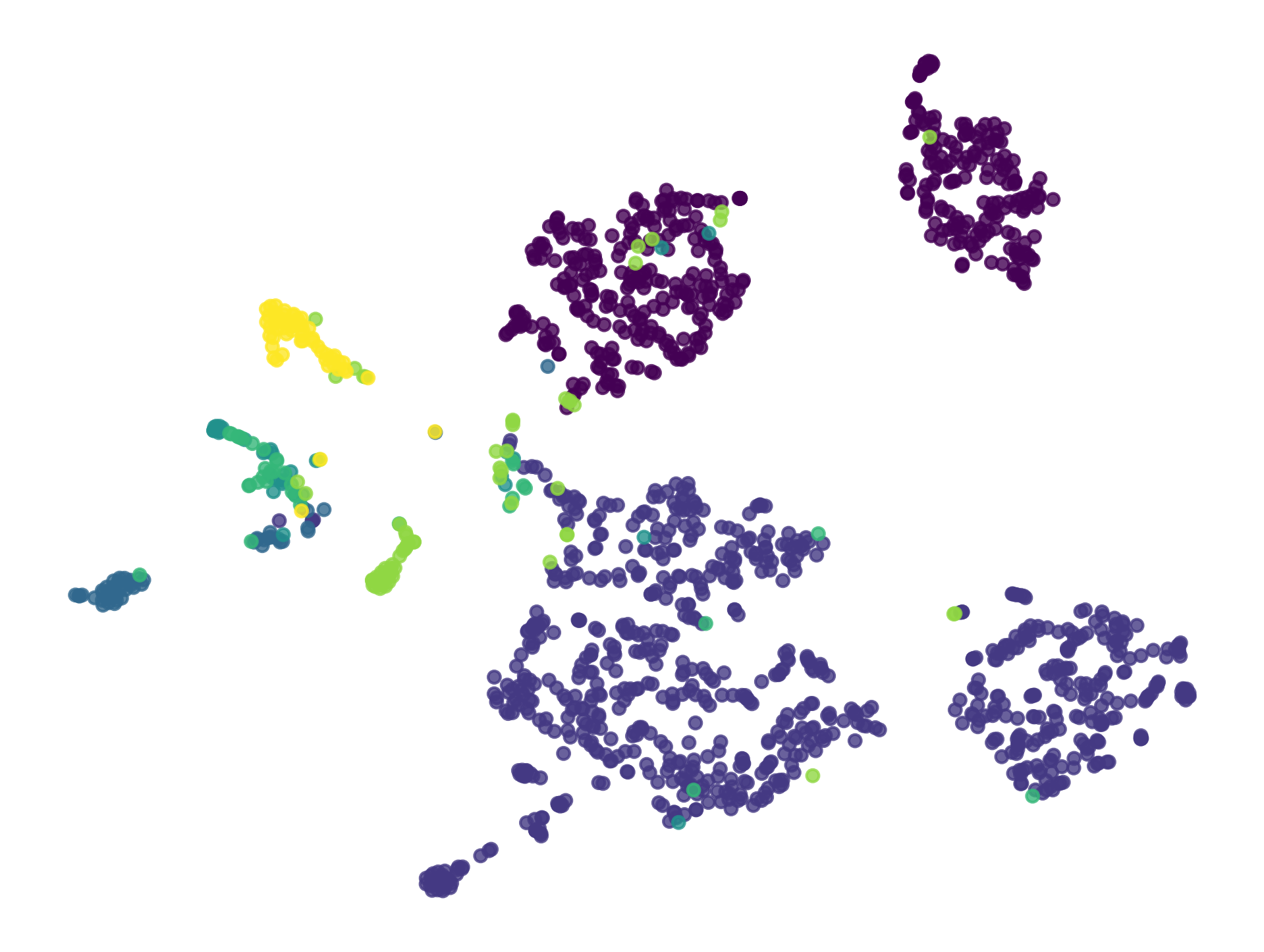}}} 
  \subfigure[Our MCFCN]{\fbox{\includegraphics[height=1.5in, width=1.7in]{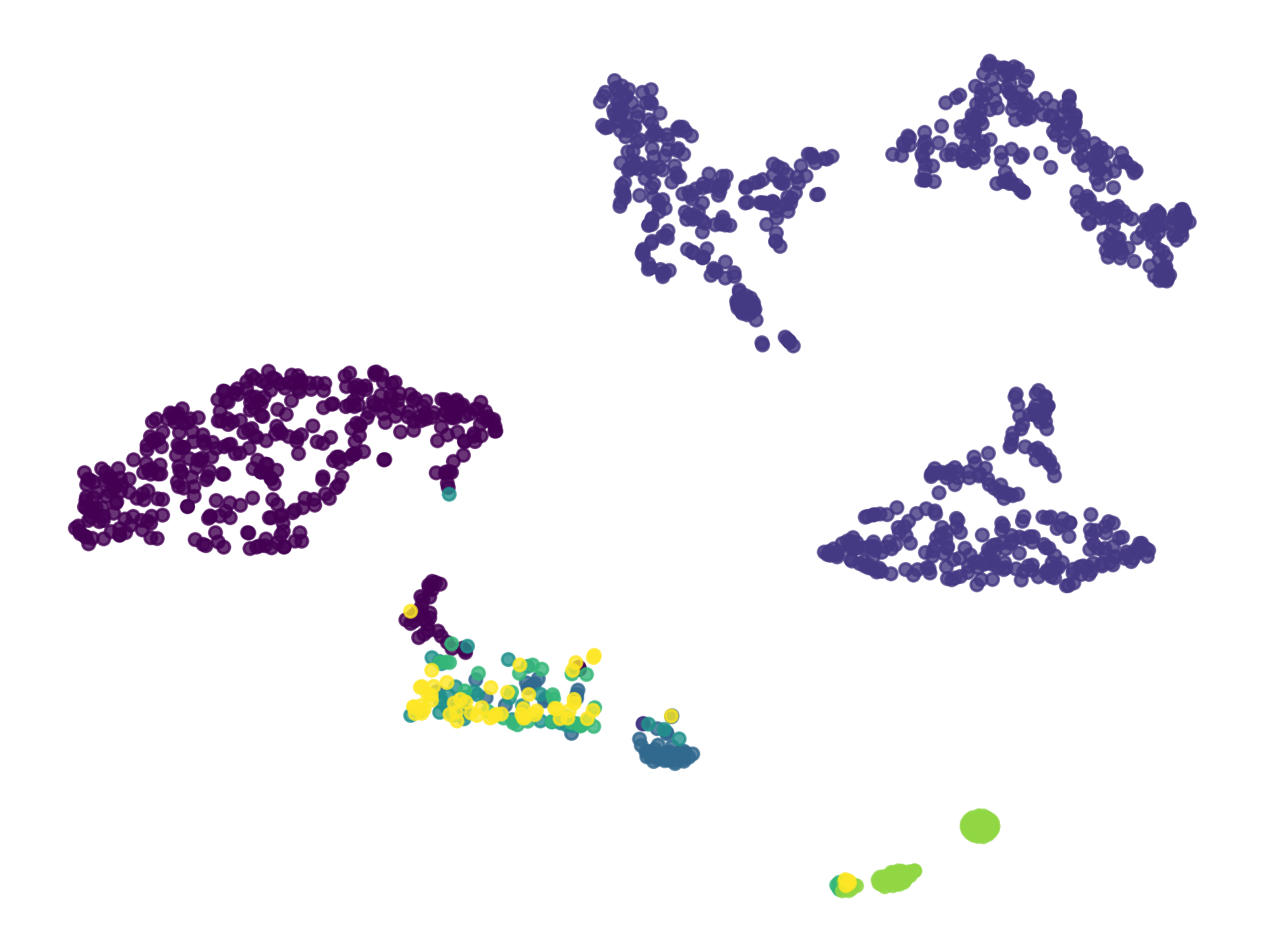}}} 
  \caption{ The t-SNE visualization of the raw features and the learned representation by different methods on the Caltech101-7.}
  \label{cal7_tSNE}
\end{figure*}

In addition, we visualized the original features and the consensus representations learned by different methods on the 3Sources, 100Leaves, and Caltech101-7 datasets using the t-SNE algorithm, where different colors were used to represent the predicted clusters. As shown in Figures \ref{3c_tSNE}, \ref{100_tSNE} and \ref{cal7_tSNE}, in all three datasets, our method can clearly represent the clustering structure of data samples. Compared with the other two methods, the consensus representation obtained by the MCFCN method is more compact among samples of the same class and has a larger distance between different classes. This indicates that our method can better learn discriminative feature representations.

\subsection{Ablation Study}
To test the effectiveness of the key techniques of the proposed method in terms of clustering performance, we conduct an ablation study. We perform experiments on the three datasets, 3Sources, BBCSport and Caltech101-7, to verify the effectiveness of the three main techniques, namely the Unified Graph Structure Adapter (UGA see Eq.\eqref{eq:3},\eqref{eq:4},\eqref{eq:5}), the Similarity Matrix Alignment Loss (SMAL see Eq.\eqref{eq:20}),  the Feature Representation Alignment Loss (FRAL see Eq.\eqref{eq:22}), and the Graph Autoencoder loss (\(L_a\) see Eq.\eqref{eq:24}). In UGA, the constructed fused graph \(\mathbf{A}_f\) is used in conjunction with the Multi-view Deep Kernel K-means Loss and Spectral Clustering Loss to obtain the final feature representation \(\mathbf{H}\). In SMAL, the Similarity Matrix Alignment Loss encourages \(\mathbf{HH}^T\) to align with both \(\mathbf{F}_v\boldsymbol{F}_v^T\) and the fused similarity matrix \(\mathbf{S}_f\). In FRAL, the Feature Representation Alignment Loss promotes the alignment between the similarity matrix \(\mathbf{S}^v\) of original features and that of linearly transformed features.  Under the premise of constructing the fused graph \(\mathbf{A}_f\), the Graph Autoencoder loss  \(L_a\) is used to make \(\mathbf{HH}^T\) as close as possible to \(\mathbf{A}_f\). The baseline model of the experiment adopts an architecture that constructs a static graph based on raw features. This model directly obtains fused features by concatenating the original features, calculates the adjacency matrix of the original features of each view through the KNN algorithm, then sums up all the view adjacency matrices and takes the average to get the average adjacency matrix. Meanwhile, it uses the same GCN architecture as MCFCN, with both the fused features and the average adjacency matrix serving as inputs. Subsequently, we gradually introduce other key components to observe changes in model performance.

Tables \ref{table:3Sources}, \ref{table:BBCSport}, and \ref{table:Caltexh101-7} present the results of ablation experiments conducted on the 3Sources, BBCSport, and Caltech101-7 datasets. From these results, we can observe that, generally speaking, as each component is applied, the performance of the model gradually improves on the three datasets. It should be noted that when only using the UGA, since the similarity matrix is calculated from the features after linear transformation rather than the original features, this causes the similarity matrix to fail to properly represent the structural information of the original features, thereby affecting the model's performance. After incorporating the SMAL, the model's performance is significantly enhanced, which indicates that SMAL can enable the model to effectively learn the structural features common to each view and align them to a unified feature representation. By adding the FRAL, the model performance is further improved, indicating that FRAL can enable the model to obtain better feature representations. Finally, the complete MCFCN model is obtained by adding \(L_a\), achieving the optimal performance.

\begin{table}[ht]
\caption{Results of different model variants on the 3Sources dataset.}
\centering
\resizebox{0.45\textwidth}{!}{
\begin{tabular}{ccccccccc}
\toprule
\multicolumn{4}{c|}{Component} & \multicolumn{5}{c}{Metrics} \\
\cmidrule(lr){1-4} \cmidrule(lr){5-8}
UGA & SMAL & FRAL & \(L_a\) & ACC & NMI & ARI & F1 \\
\midrule
& & & & 0.5917 & 0.5966 & 0.4101 & 0.5282 \\
\checkmark & & & & 0.6449 & 0.6796 & 0.5299 & 0.5748 & \\
\checkmark & \checkmark & & & 0.8106 & 0.6816 & 0.6458 & 0.7851 \\
\checkmark & \checkmark & \checkmark & & 0.8382 & 0.7092 & 0.6648 & 0.7894 \\
\checkmark & \checkmark & \checkmark & \checkmark & \textbf{0.8402} & \textbf{0.7406} & \textbf{0.6796} & \textbf{0.8342} \\
\bottomrule
\end{tabular}
}
\label{table:3Sources}
\end{table}

\begin{table}[ht]
\caption{Results of different model variants on the BBCSport dataset.}
\centering
\resizebox{0.45\textwidth}{!}{
\begin{tabular}{ccccccccc}
\toprule
\multicolumn{4}{c|}{Component} & \multicolumn{5}{c}{Metrics} \\
\cmidrule(lr){1-4} \cmidrule(lr){5-8}
UGA & SMAL & FRAL & \(L_a\) & ACC & NMI & ARI & F1 \\
\midrule
& & & & 0.7205 & 0.7512 & 0.6141 & 0.6862 \\
\checkmark & & & &0.6948 & 0.5963 & 0.5416 & 0.6690 & \\
\checkmark & \checkmark & & & 0.8860 & 0.7658 & 0.8128 & 0.8358 \\
\checkmark & \checkmark & \checkmark & & 0.9540 & 0.8708 & 0.8859 & 0.9504 \\
\checkmark & \checkmark & \checkmark & \checkmark &  \textbf{0.9651} & \textbf{0.8964} & \textbf{0.9108} & \textbf{0.9632} \\
\bottomrule
\end{tabular}
}
\label{table:BBCSport}
\end{table}

\begin{table}[h!]
\caption{Results of different model variants on the Caltech101-7 dataset.}
\centering
\resizebox{0.45\textwidth}{!}{
\begin{tabular}{ccccccccc}
\toprule
\multicolumn{4}{c|}{Component} & \multicolumn{5}{c}{Metrics} \\
\cmidrule(lr){1-4} \cmidrule(lr){5-8}
UGA & SMAL & FRAL & \(L_a\) & ACC & NMI & ARI & F1 \\
\midrule
& & & & 0.5800 & 0.5941 & 0.4812 & 0.38077 \\
\checkmark & & & & 0.5739 & 0.5900 & 0.4517 & 0.3652 & \\
\checkmark & \checkmark & & & 0.7014 & 0.6399 & 0.5423 & 0.5446 \\
\checkmark & \checkmark & \checkmark & & 0.8639 & 0.7291 & 0.8193 & 0.6559 \\
\checkmark & \checkmark & \checkmark & \checkmark & \textbf{0.8881} & \textbf{0.7596} & \textbf{0.8461} & \textbf{0.6616} \\
\bottomrule
\end{tabular}
}
\label{table:Caltexh101-7}
\end{table}

 \begin{figure*}[h!]
  \centering
  \subfigure[]{\includegraphics[height=1.35in]{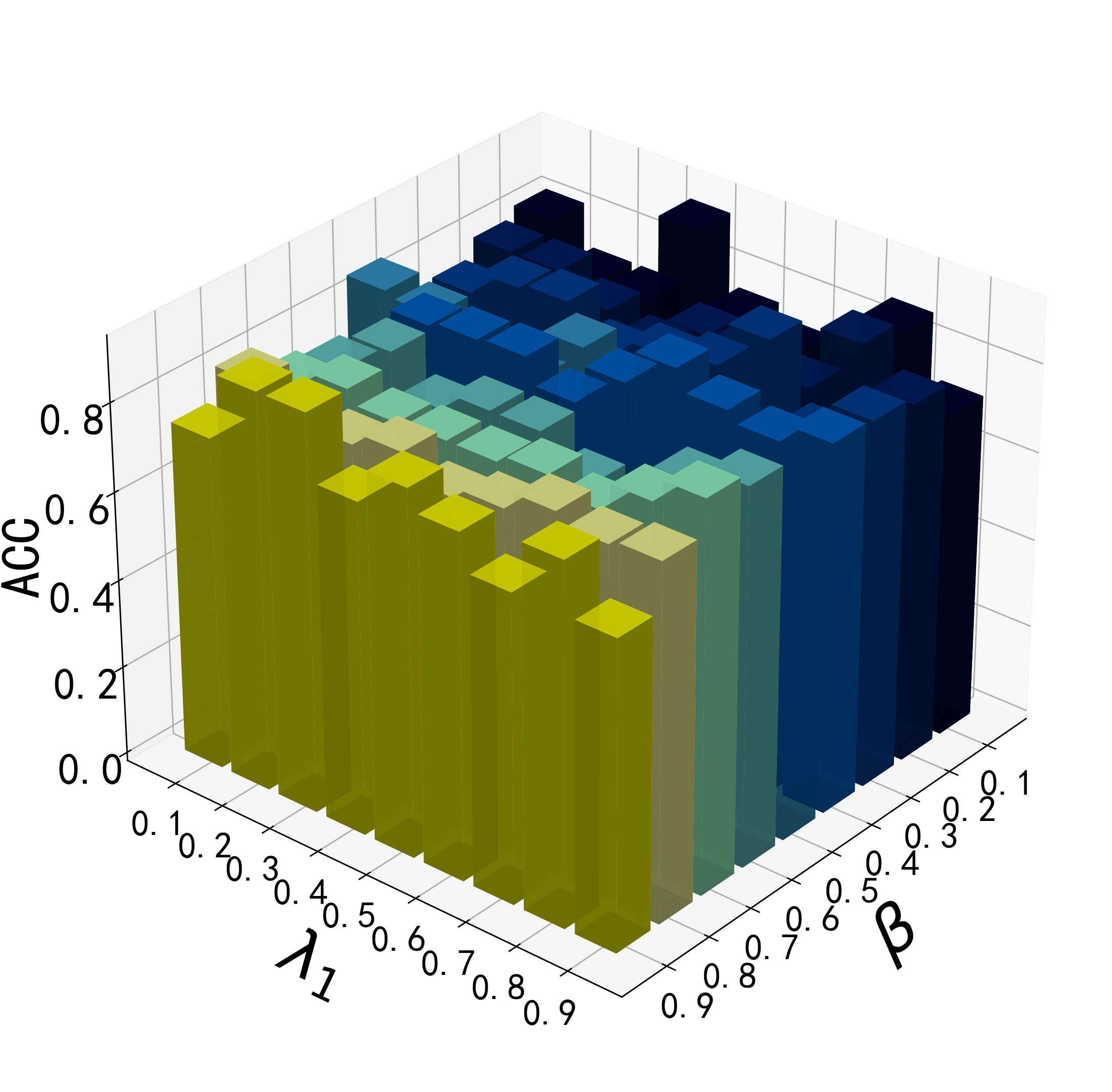}}
  \subfigure[]{\includegraphics[height=1.35in]{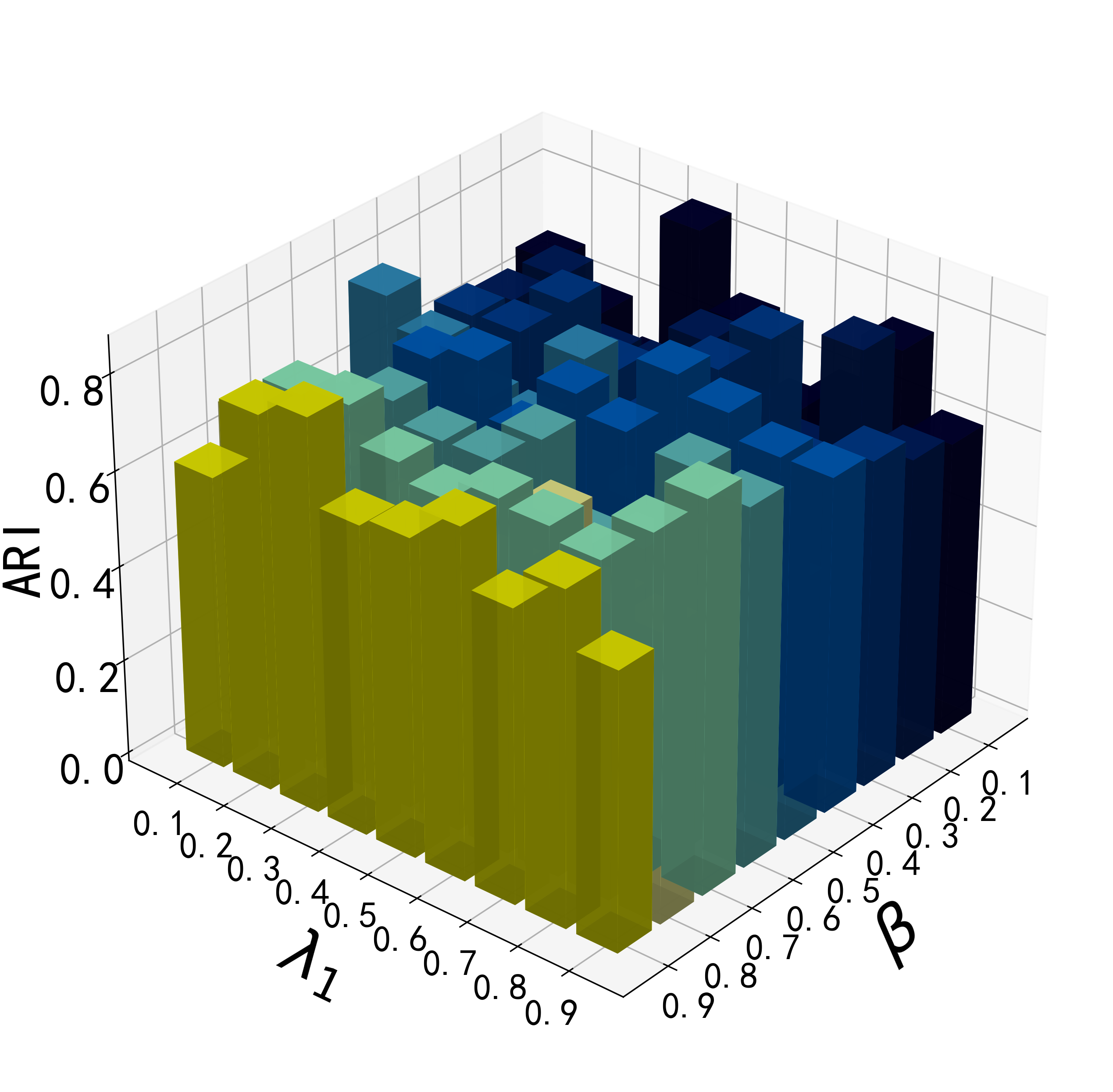}}
  \subfigure[]{\includegraphics[height=1.35in]{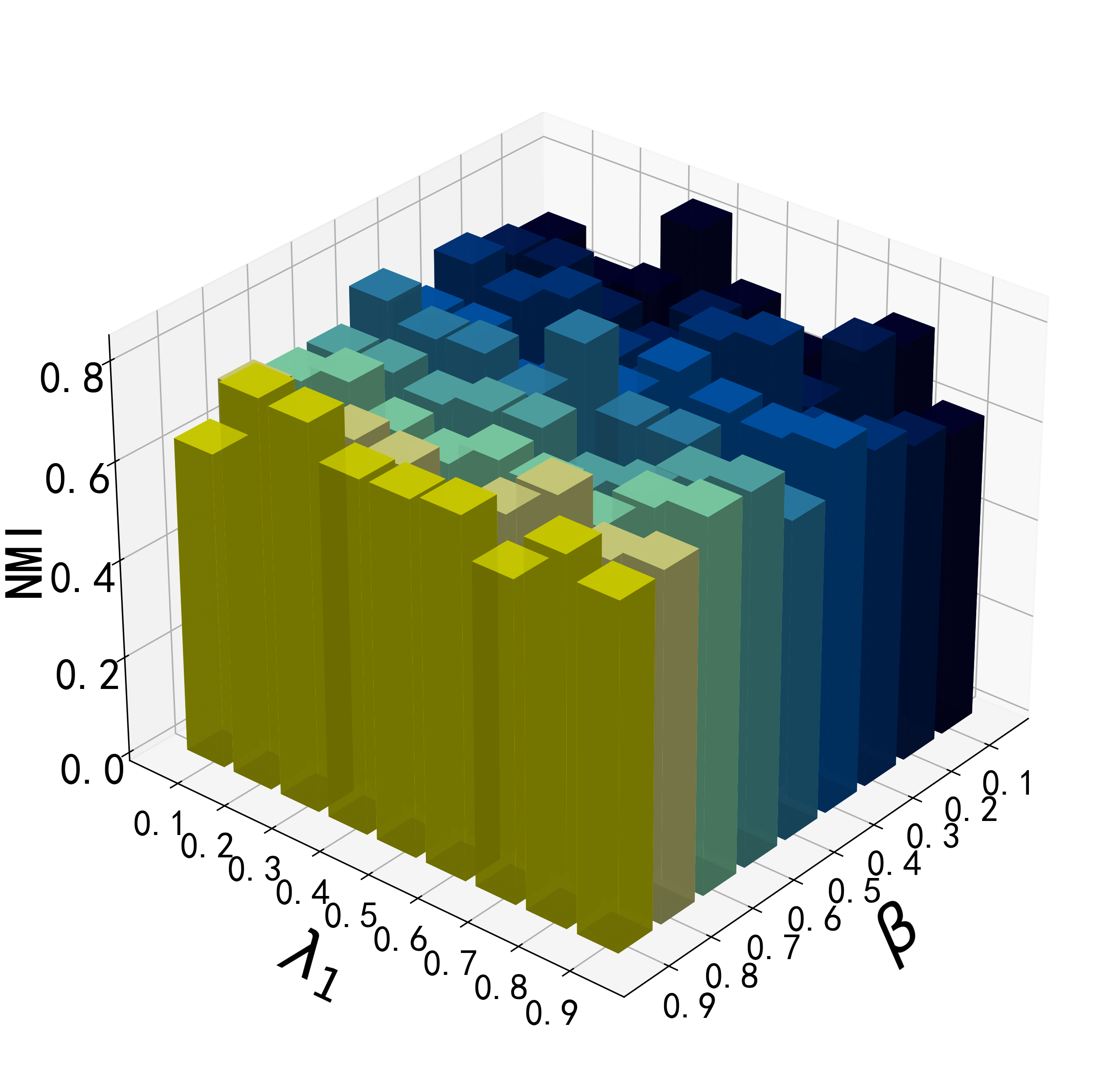}}
  \subfigure[]{\includegraphics[height=1.35in]{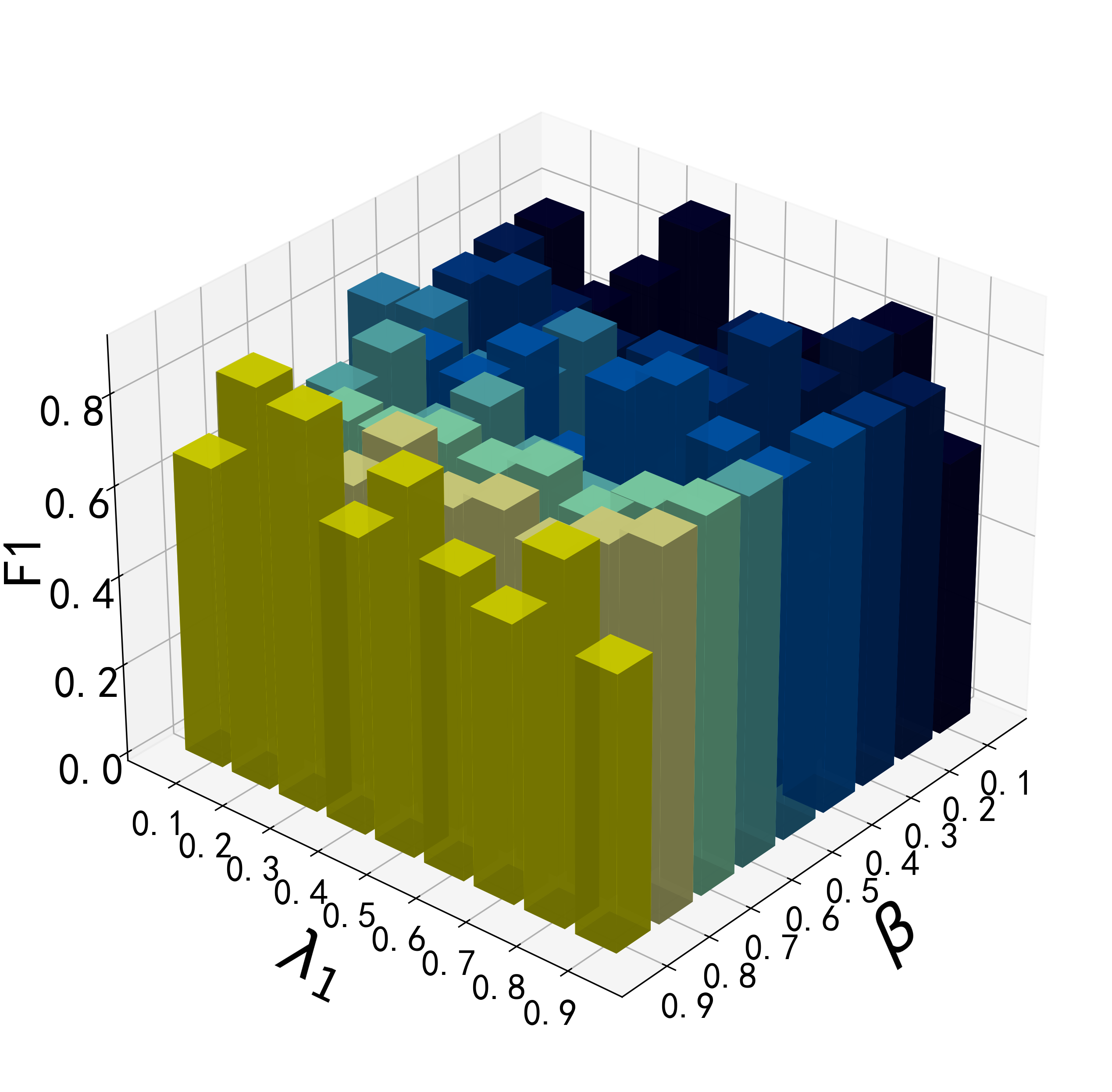}}
  \subfigure[]{\includegraphics[height=1.35in]{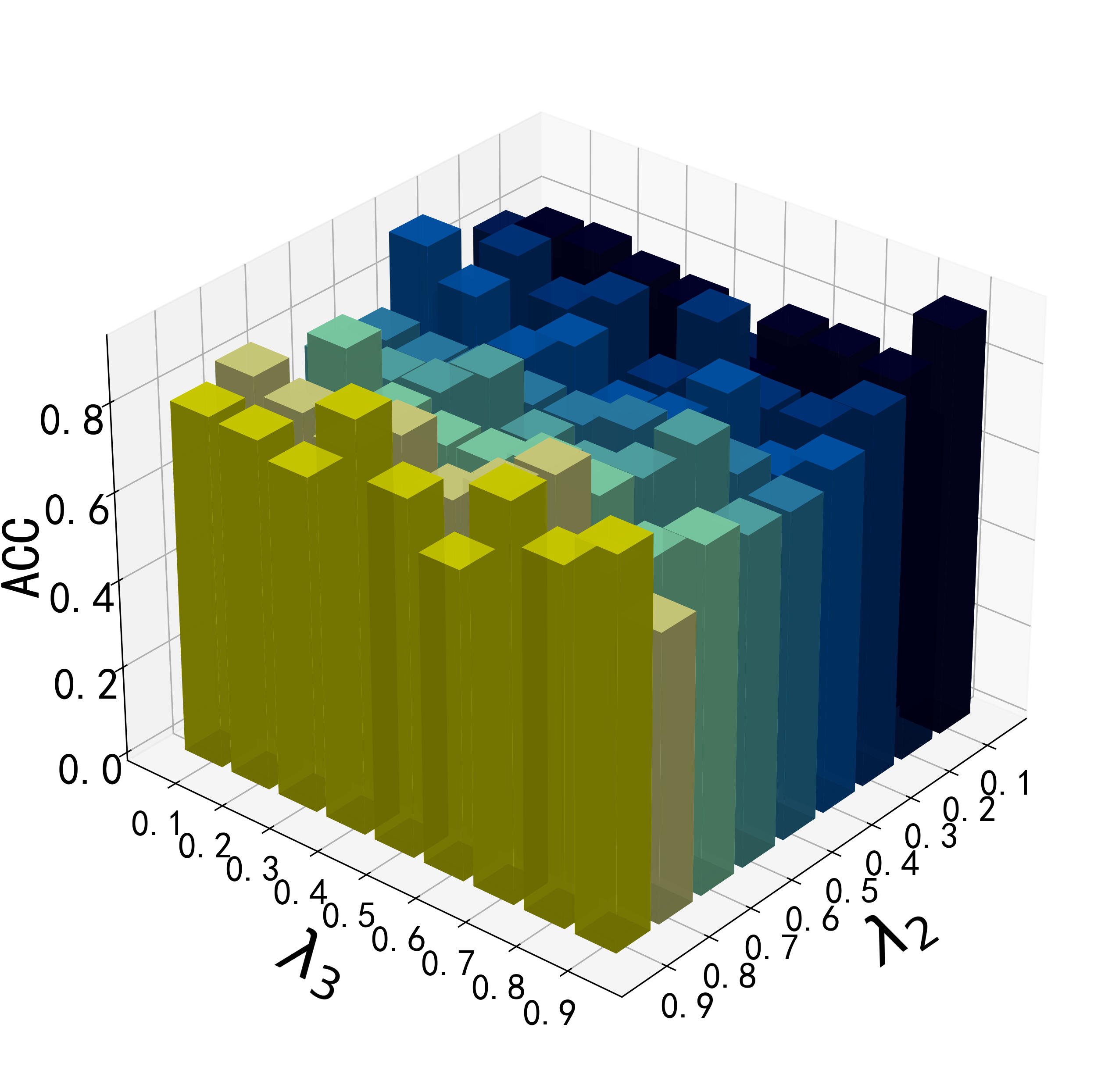}}
  \subfigure[]{\includegraphics[height=1.35in]{3s_acc_2.png}} 
  \subfigure[]{\includegraphics[height=1.35in]{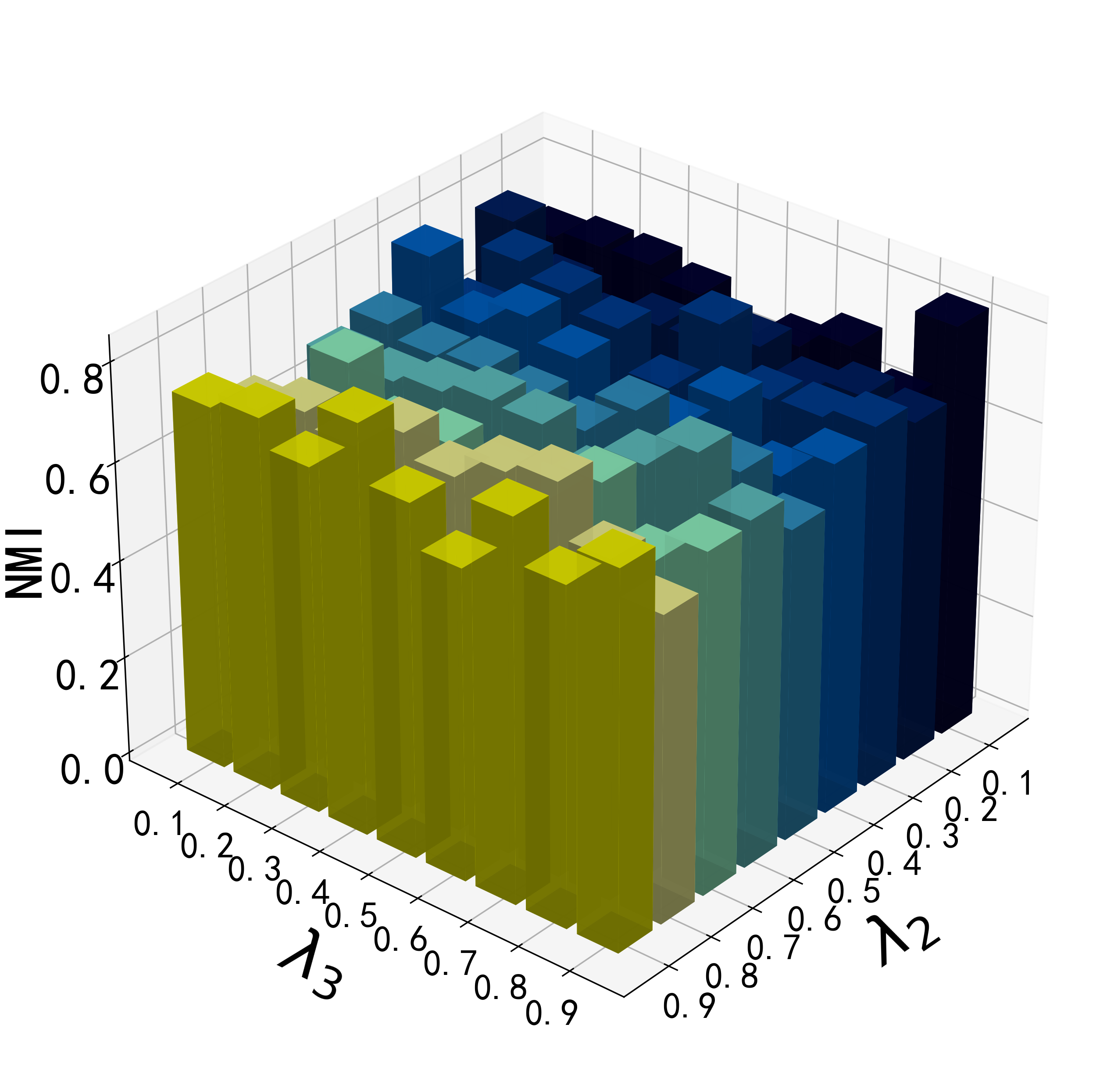}}
  \subfigure[]{\includegraphics[height=1.35in]{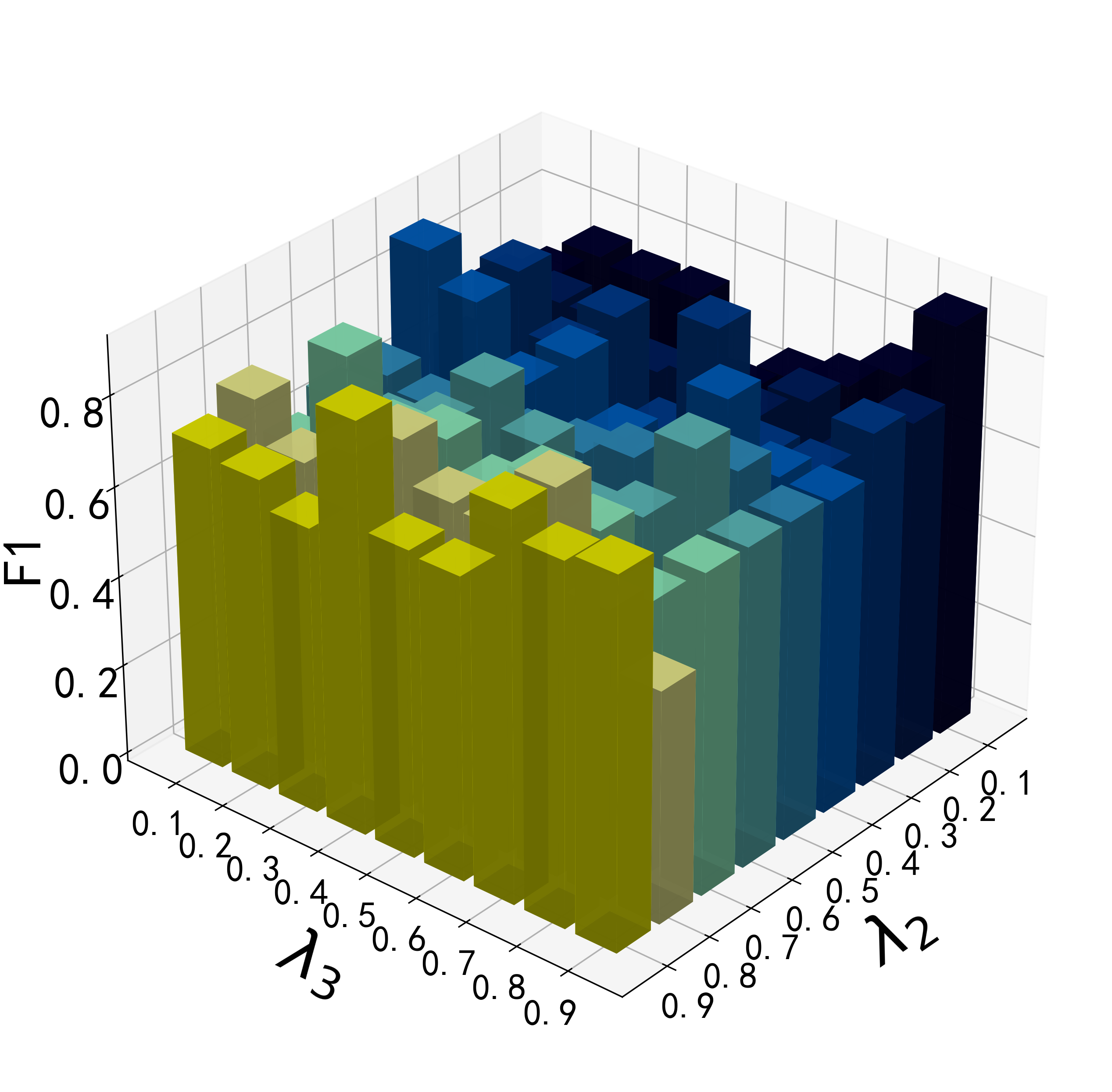}}
  \caption{Parameter sensitivity analysis of $\beta$ vs. $\lambda_1$ and $\lambda_2$ vs. $\lambda_3$ on the 3Sources dataset measured  by the ACC, NMI, ARI and F1 metrics.}
  \label{3s_3d}
\end{figure*}

\subsection{Parameters Analysis}
\begin{figure*}
  \centering
  \subfigure[]{\includegraphics[height=1.35in]{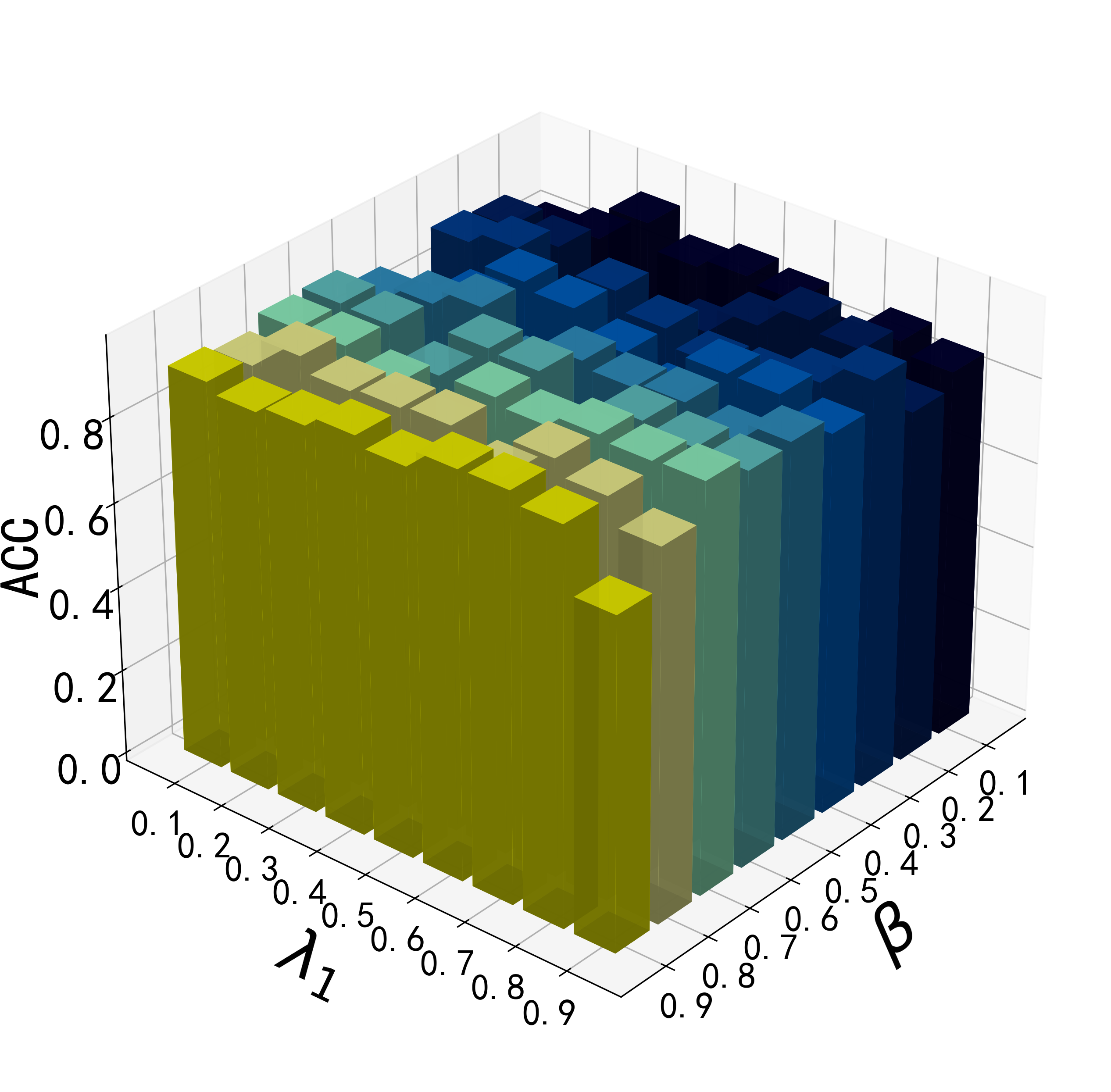}}
  \subfigure[]{\includegraphics[height=1.35in]{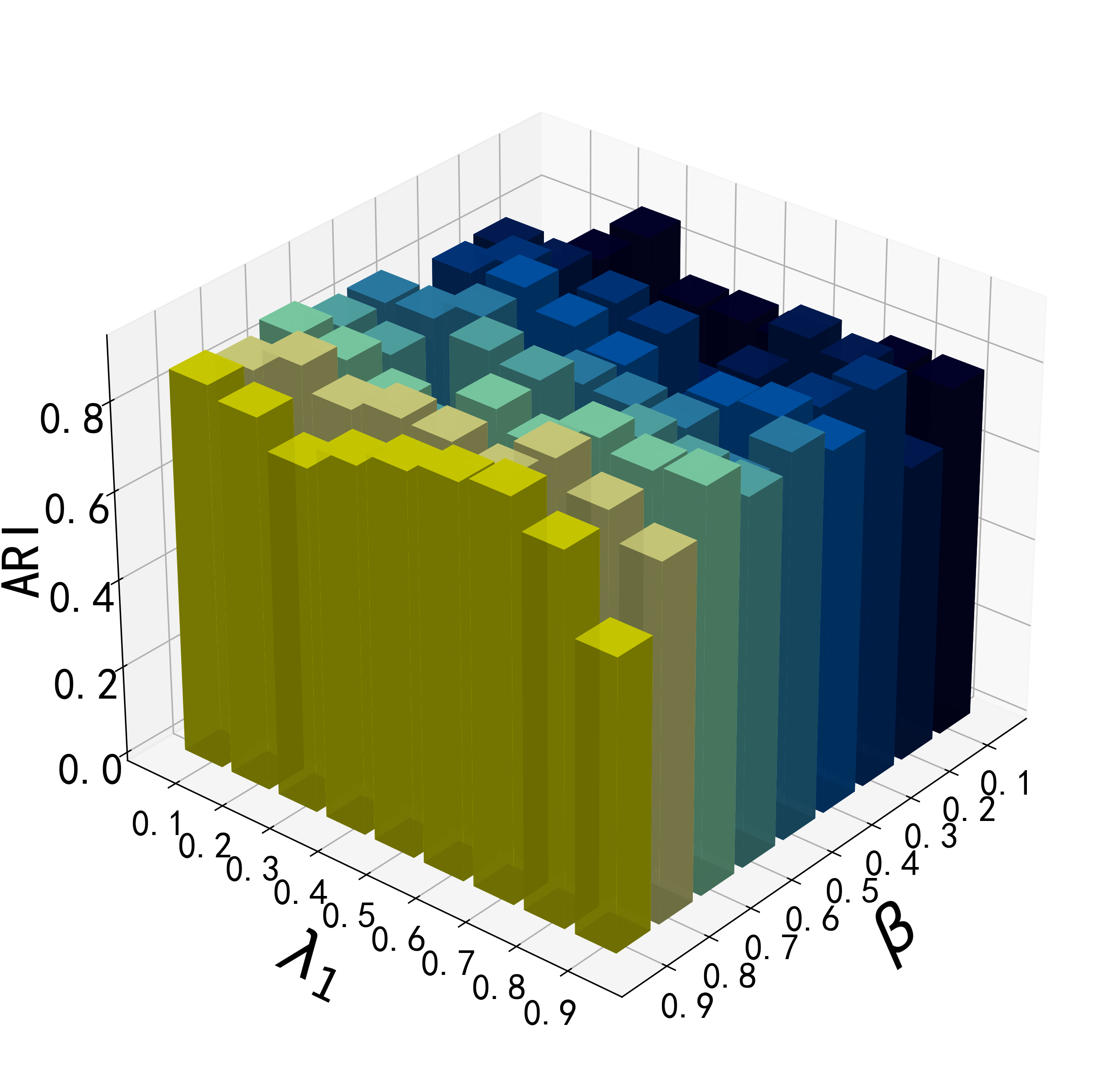}}
  \subfigure[]{\includegraphics[height=1.35in]{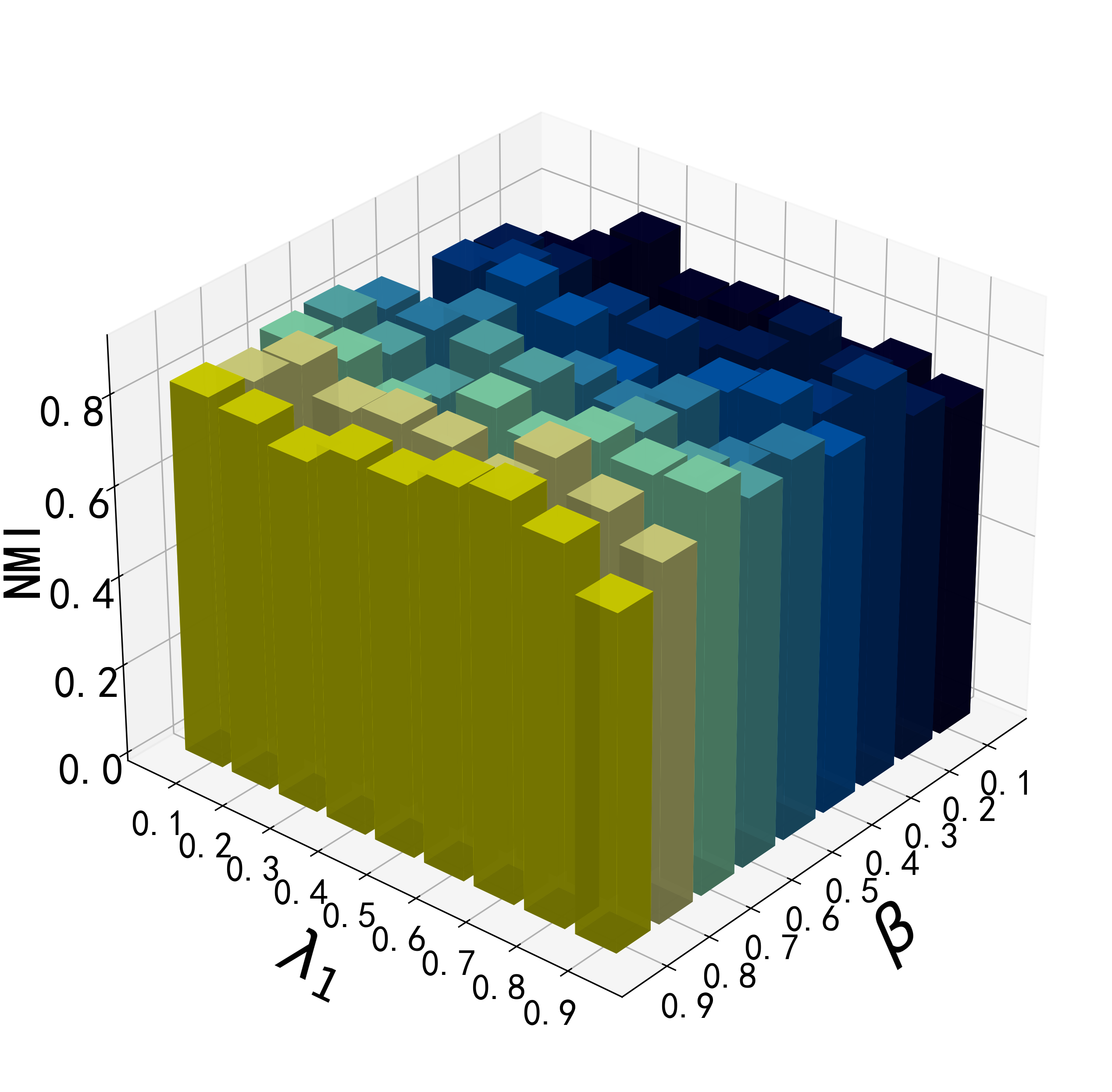}}
  \subfigure[]{\includegraphics[height=1.35in]{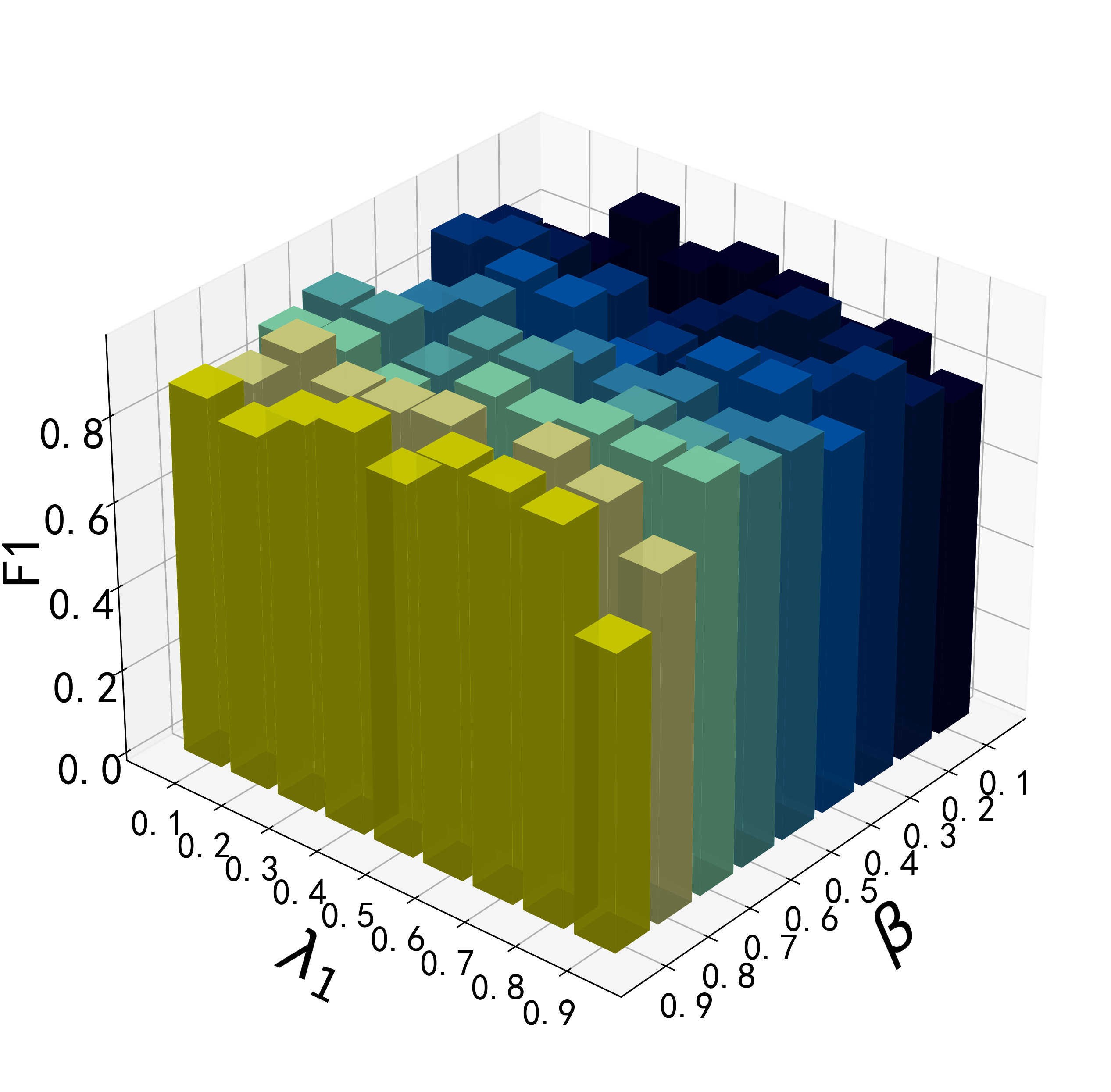}}
  \subfigure[]{\includegraphics[height=1.35in]{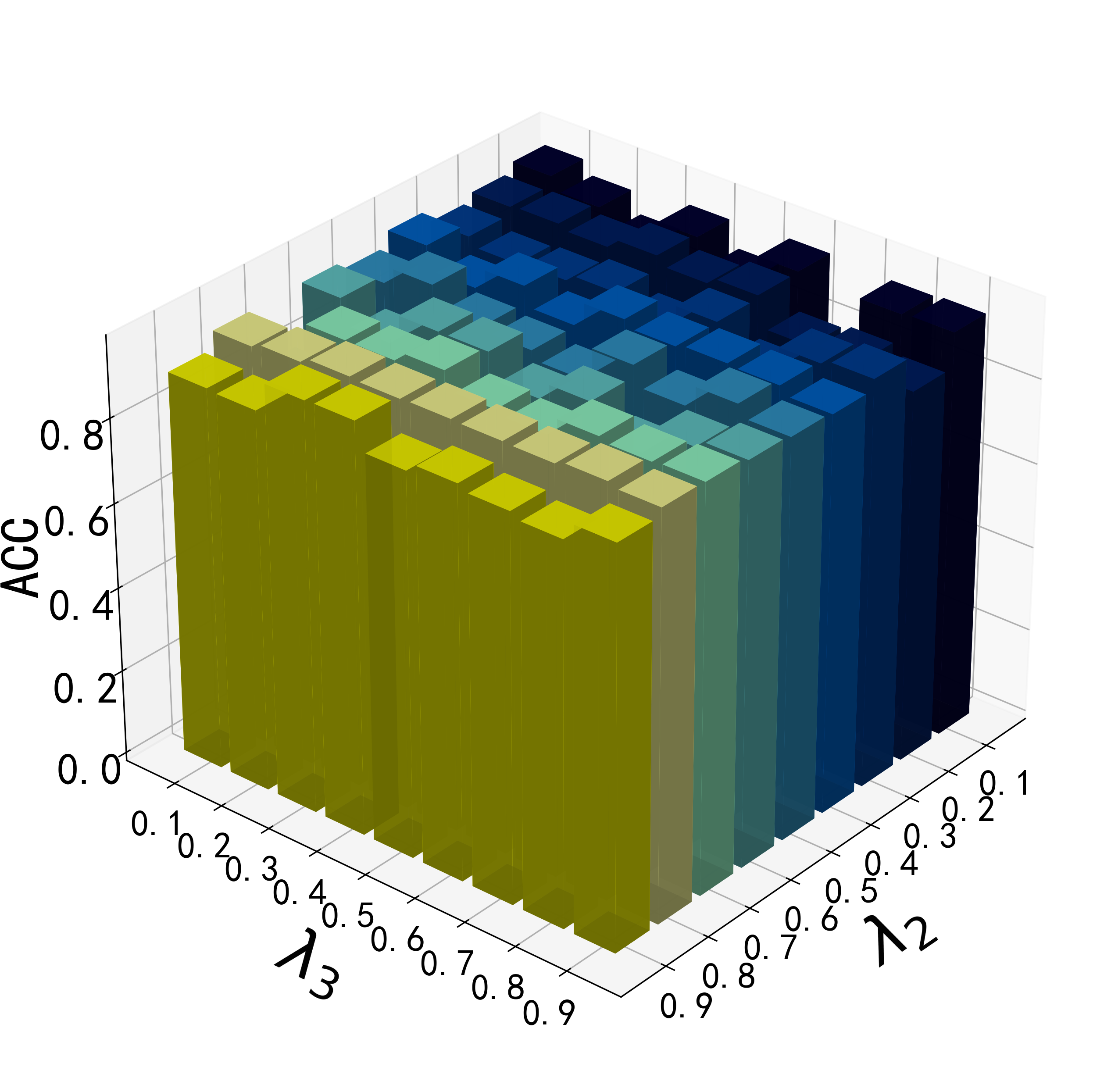}}
  \subfigure[]{\includegraphics[height=1.35in]{bbc_acc_2.png}} 
  \subfigure[]{\includegraphics[height=1.35in]{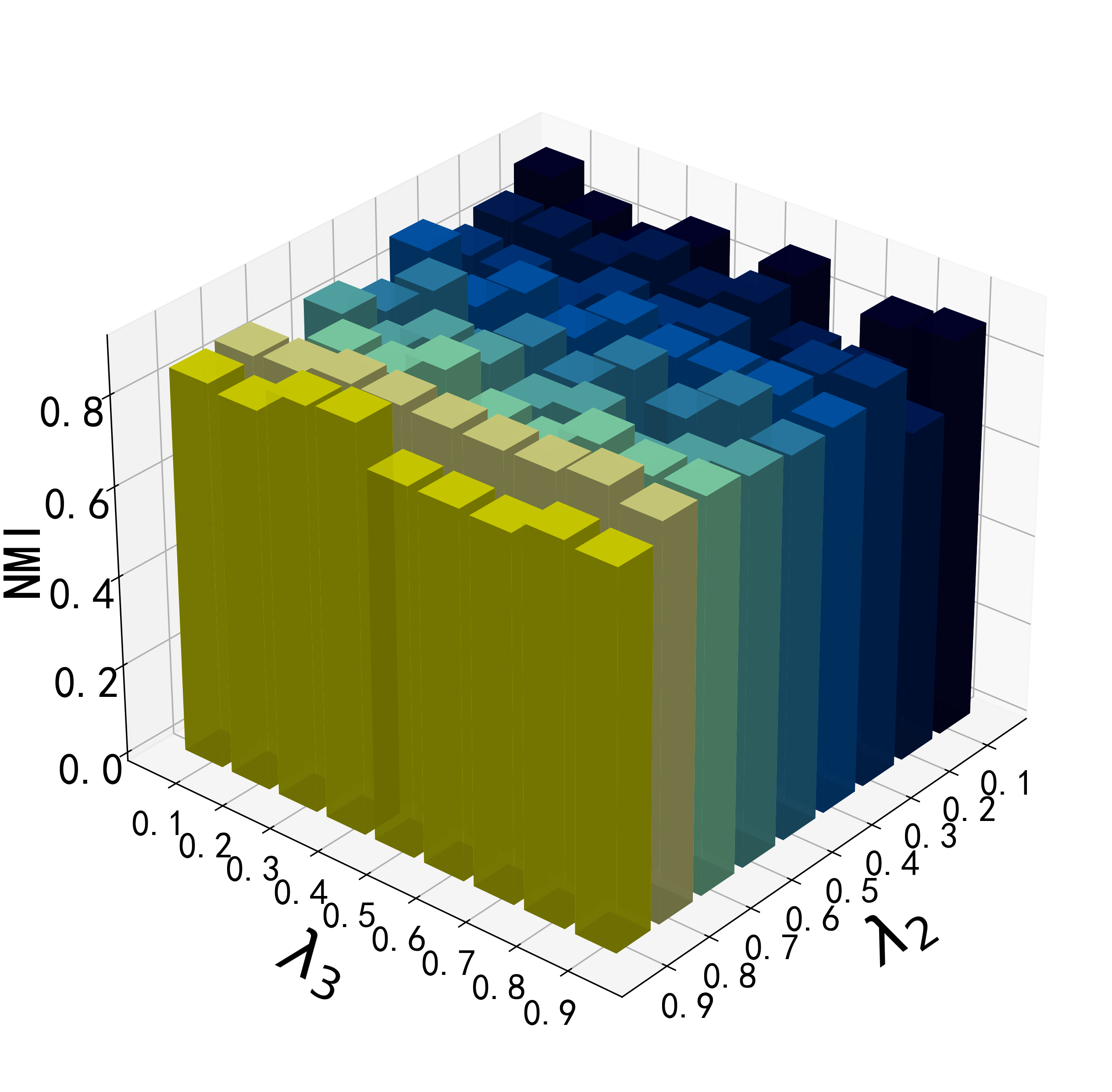}}
  \subfigure[]{\includegraphics[height=1.35in]{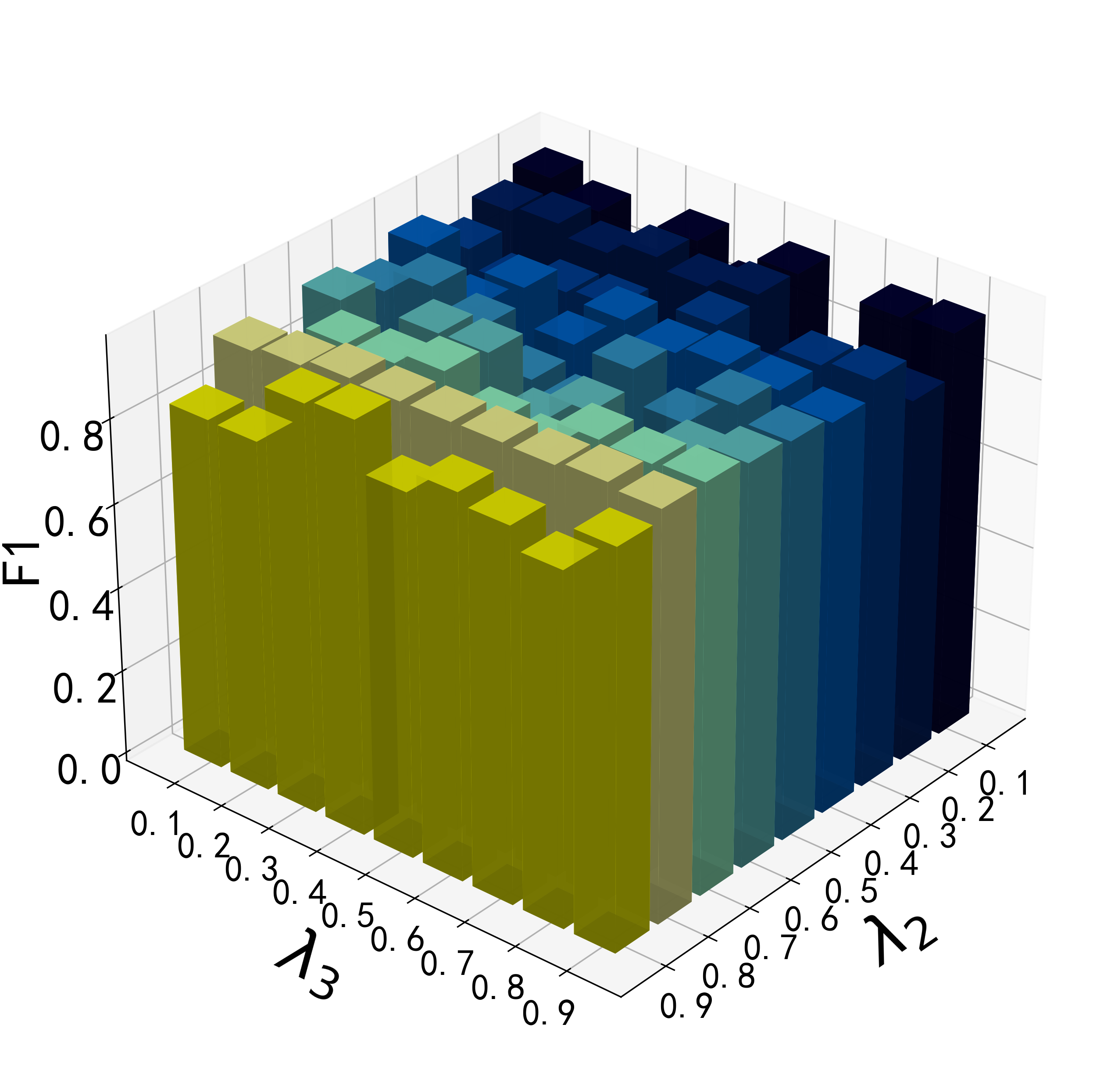}}
  \caption{Parameter sensitivity analysis of $\beta$ vs. $\lambda_1$ and $\lambda_2$ vs. $\lambda_3$ on the BBCSport dataset measured  by the ACC, NMI, ARI and F1 metrics.}
  \label{bbc_3d}
\end{figure*}

\begin{figure*}[h!]
  \setlength{\fboxsep}{0pt} 
  \setlength{\fboxrule}{0.1pt} 
  \centering
  \subfigure[3Sources]{\fbox{\includegraphics[height= 2in, width=3.6in]{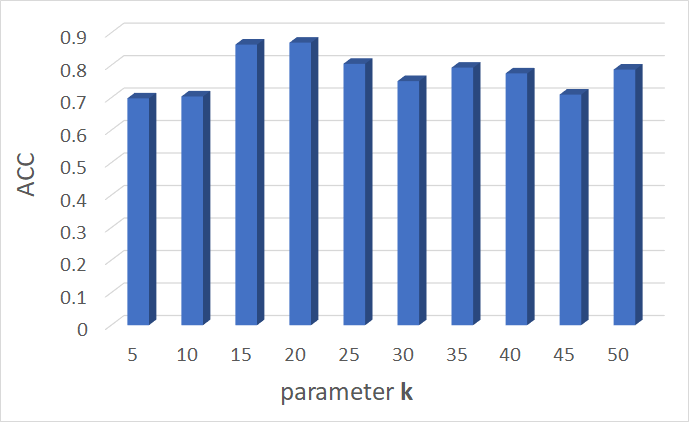}}}
  \subfigure[BBCSport]{\fbox{\includegraphics[ height= 2in, width=3.6in]{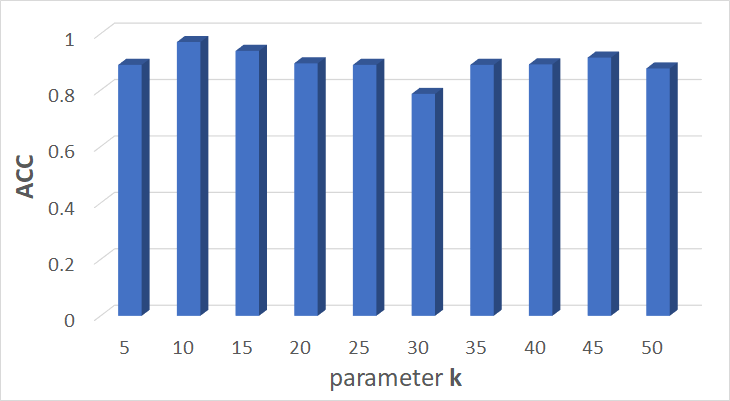}}} 
  \caption{The performance of the proposed method is evaluated across different $k$ values on the 3Sources and BBCSport datasets. }
  \label{3s_bbc_k}
\end{figure*}

In our MCFCN model, the weight parameters $\beta$, $\lambda_1$, $\lambda_2$, $\lambda_3$, and the $k$-value in Graph Sparsification are five relatively important hyperparameters. In this section, we select ACC, NMI, ARI, and F1 as evaluation metrics, and study and analyze the sensitivities of these hyperparameters on the 3Sources and BBCSport datasets. We present the experimental results in Figure \ref{3s_3d}, Figure \ref{bbc_3d}, and Figure \ref{3s_bbc_k}, respectively.

\textbf{Parameter Sensitivities of $\beta$, $\lambda_1$, $\lambda_2$ and $\lambda_3$.} We conduct extensive experiments with different combinations of parameters $\beta$, $\lambda_1$, $\lambda_2$ and $\lambda_3$. within the range of [0.1, 0.2, 0.3, 0.4, 0.5, 0.6, 0.7, 0.8, 0.9]. As shown in Figure \ref{3s_3d} (a-d) and Figure \ref{bbc_3d} (a-d), parameters $\lambda_2$ and $\lambda_3$ exhibit relatively stable performance within the set range. Parameters $\beta$, $\lambda_1$ show more sensitive characteristics on the 3Sources dataset. Therefore, according to the experimental results, we can select candidate parameters $\beta$, $\lambda_1$ for different datasets.

\textbf{Parameter Sensitivity of $k$.} In Figure \ref{3s_bbc_k}, we set the range of $k$ as [5, 10, 15, 20, 25, 30, 35, 40, 45]. It can be observed that when the value of $k$ is small (e.g., 5 or 10), it may lead to insufficient feature structure information. Conversely, when the value of $k$ is too large, excessive noise may be introduced. Therefore, we need to select different candidate values of $k$ for different datasets.

\section{Conclusion}
\label{sec:conclusion}
In this work, we propose MCFCN, a Multi-View Fusion Graph Convolutional Network for multi-view clustering. Unlike existing methods, we obtain a unified fused representation of multi-view features and a consensus graph structure through the Multi-View Feature Fusion Module and Unified Graph Structure Adapter, and integrate them with Graph Convolutional Networks (GCN) into a single network to achieve clustering-oriented joint optimization. Extensive experiments and ablation studies on multiple benchmark datasets demonstrate the effectiveness and superiority of MCFCN. For future work, we plan to extend the proposed method with inter-view contrastive strategies to achieve better performance. Additionally, considering the common occurrence of partial view missing in real-world scenarios, another research direction is to generalize the proposed method to incomplete multi-view clustering tasks.


\end{document}